\documentclass{article}

\usepackage[utf8]{inputenc}    
\usepackage[T1]{fontenc}       
\usepackage{CJKutf8}           

\usepackage{amsmath, amssymb, amsfonts}
\usepackage{xfrac}
\usepackage{microtype}         

\usepackage{booktabs}          
\usepackage{multirow}          
\usepackage{makecell}          
\usepackage{threeparttable}    
\usepackage{tablefootnote}     
\usepackage{tabularx}          

\usepackage{graphicx}

\usepackage{xcolor}
\usepackage{textcomp}

\usepackage[colorlinks=true, allcolors=blue]{hyperref}
\usepackage{url}
\usepackage{doi}
\usepackage{natbib}
\bibpunct{[}{]}{;}{a}{,}{,}     

\usepackage{arxiv}

\title{LLM-BT-Terms: Back-Translation as a Framework for Terminology Standardization and Dynamic Semantic Embedding}

\author{ \href{https://orcid.org/0000-0003-1826-1850}{\includegraphics[scale=0.06]{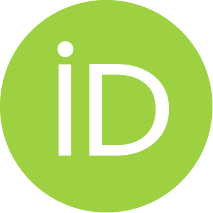}\hspace{1mm}Li Weigang} \\
	TransLab, Computer Science Department\\
	University of  Brasilia\\
        Brasília, Brazil \\
	\texttt{weigang@unb.br} \\
	\And
	\href{https://orcid.org/0000-0002-1288-7695}{\includegraphics[scale=0.06]{orcid.pdf}\hspace{1mm}Pedro Carvalho Brom} \\
	Math Department\\
	Federal Institute of Brasilia\\
	Brasília, Brazil \\
	\texttt{pedro.brom@ifb.edu.br} \\
}

\renewcommand{\shorttitle}{LLM-BT-Terms for Terminology Standardization and Dynamic Semantic Embedding}

\hypersetup{
pdftitle={LLM-BT-Terms: Back-Translation as a Framework for Terminology Standardization and Dynamic Semantic Embedding},
pdfsubject={cs.CL},
pdfauthor={Li Weigang, Pedro Carvalho Brom},
pdfkeywords={Back-translation, Explainable AI, LLM, Dynamic semantic embedding, Terminology extraction, Terminology standardization},
}

\begin{document}
\maketitle

\begin{abstract}
The rapid growth of English technical terms challenges traditional expert-driven standardization, especially in fast-evolving fields like AI and quantum computing. Manual methods struggle to ensure multilingual consistency. We propose \textbf{LLM-BT}, a back-translation framework powered by large language models (LLMs) to automate terminology verification and standardization via cross-lingual semantic alignment. Our contributions are: \textbf{(1) Term-Level Consistency Validation:} Using English $\rightarrow$ intermediate language $\rightarrow$ English back-translation, LLM-BT achieves high term consistency across models (e.g., GPT-4, DeepSeek, Grok), with case studies showing over 90\% exact or semantic matches. \textbf{(2) Multi-Path Verification Workflow:} A novel ``Retrieve--Generate--Verify--Optimize'' pipeline integrates serial (e.g., EN $\rightarrow$ ZHcn $\rightarrow$ ZHtw $\rightarrow$ EN) and parallel (e.g., EN $\rightarrow$ Chinese/Portuguese $\rightarrow$ EN) BT routes. BLEU and term accuracy indicate strong cross-lingual robustness (BLEU $>$ 0.45; Portuguese accuracy 100\%). \textbf{(3) Back-Translation as Semantic Embedding:} BT is conceptualized as dynamic semantic embedding, revealing latent meaning trajectories. Unlike static embeddings, LLM-BT provides transparent path-based embeddings shaped by model evolution. LLM-BT transforms back-translation into an active engine for multilingual terminology standardization, enabling human--AI collaboration: machines ensure semantic fidelity, humans guide cultural interpretation. This infrastructure supports terminology governance across scientific and technological fields worldwide.
\end{abstract}

\keywords{ Back-translation \and Cross-lingual Alignment \and Explainable AI \and Dynamic semantic embedding \and LLM \and Terminology extraction \and Terminology standardization}

\section{Introduction}

Scientific and technical terminology is undergoing unprecedented expansion, fueled by the rapid advancement of digital communication and emerging technologies. Estimates suggest that English vocabulary grows by 1,000 to 4,000 terms annually, surpassing one million entries, with substantial contributions from fields such as artificial intelligence and biomedicine \citep{kharkovskaya2020language}. This explosive growth presents serious challenges for terminology standardization across non-English languages. Traditional expert-driven translation and validation cycles are inefficient, often unable to keep pace with the velocity of scientific progress. In response, some scholars advocate for direct transliteration or wholesale adoption of English terms (e.g., retaining ``Transformer'' untranslated) to bypass low-efficiency standardization pipelines. However, such practices raise deeper concerns regarding linguistic autonomy: when a language is systematically excluded from scientific naming, its epistemic continuity and cultural sovereignty risk marginalization \citep{darvin2016}.

\begin{figure}[htbp]
\centering
\includegraphics[width=0.80\textwidth]{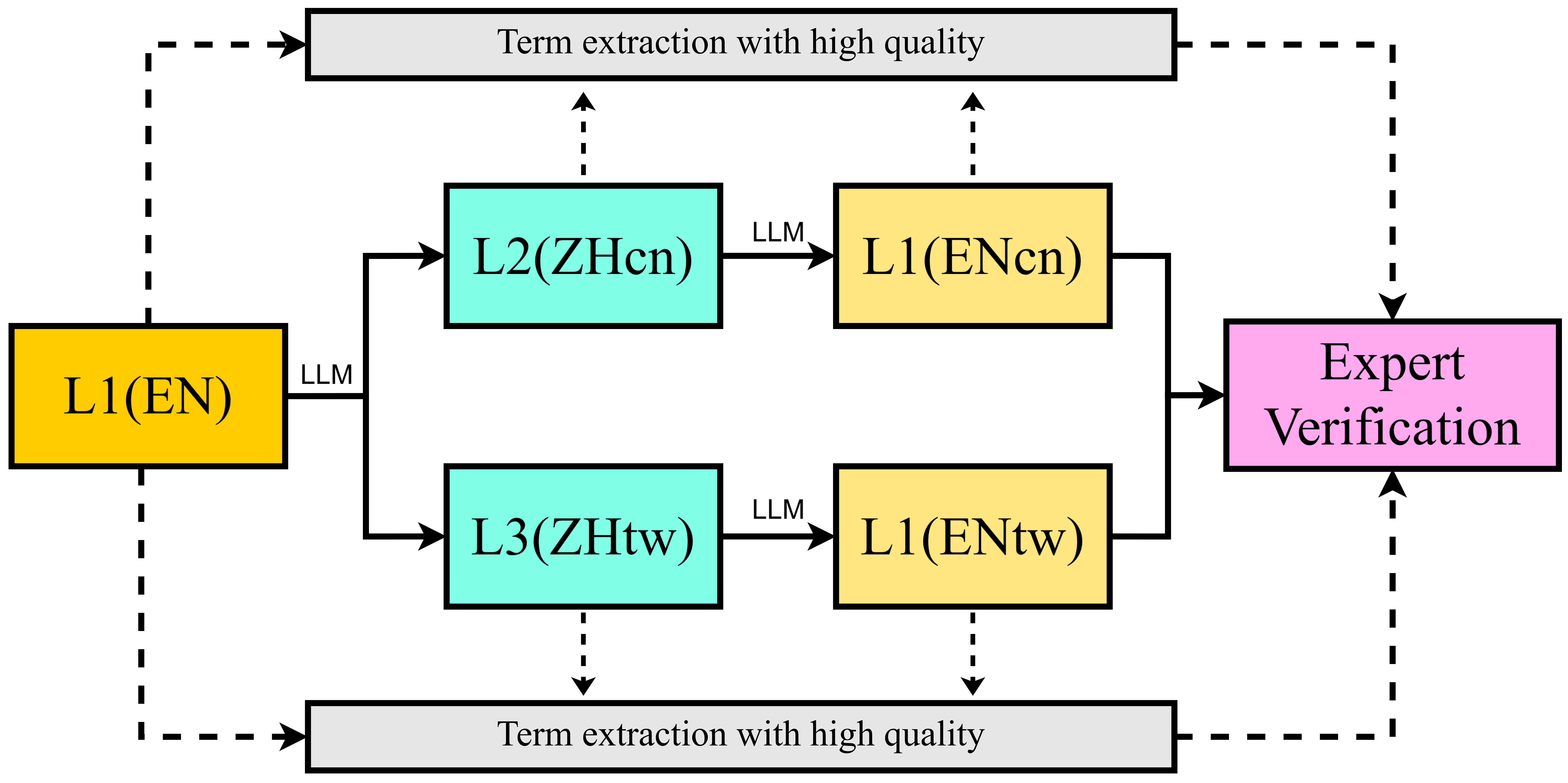}
\caption{Conceptual diagram of LLM-based back-translation for terminology standardization. $L1$(EN) is the original English input, $L2$(ZHcn)/$L3$(ZHtw) as intermediate languages are simplified/traditional Chinese translations and $L1$(ENcn)/$L1$(ENtw) are the corresponding back-translations. It shows the workflow of LLM-BT as Retrieve-Generate-Verify-Optimize to get the terms.}
\label{fig: LLM-BT-Terms}
\end{figure}

At a conceptual level, terminology serves not only as a vehicle for scientific expression but also as a core mechanism for knowledge construction and cultural continuity \citep{kageura1996methods,temmerman2000}. Languages excluded from high-tech discourse often struggle to maintain epistemological relevance and symbolic legitimacy. In this context, the development of scalable, semantically reliable and multilingual terminology generation frameworks has emerged as a critical area at the intersection of linguistic engineering, artificial intelligence and science communication.

National and international standardization bodies have acknowledged this urgency. For example, the China National Committee for Terminology in Science and Technology has established over 160 term systems across foundational sciences and applied domains through collaborative disciplinary committees \citep{fengzhiwei1997,liuqing2015,yang2023}. Nonetheless, current frameworks face significant limitations in the digital era: expert reviews typically require 12 to 18 months; update cycles are rigid and reactive; and mechanisms for capturing emerging terms from sources such as arXiv or ISO standards remain underdeveloped. In parallel, China's advancements in 5G/6G, quantum technologies and AI demand efficient mechanisms for both internal localization and external dissemination of scientific terminology \citep{cao2025}.

In Brazil, the development of archival terminology dictionaries involves collaboration among archivists, librarians, linguists, and public institutions to create a standardized technical vocabulary that reflects Brazilian Portuguese while aligning with international standards, such as those of the International Council on Archives (ICA). The Brazilian Standards Association (ABNT) and the Brazilian Archivists Association (AAB) lead terminology standardization efforts, while the National Archives and the São Paulo State Public Archives contributed to regional and national dictionaries in the 1980s and 1990s, adapting English terms to local practices \citep{camargo2005dicionario}. In 1997, the Ibero-American Group for Administrative Archives Processing fostered term convergence among Brazil, Portugal, and Spanish-speaking countries, influencing English translations. Terms without direct equivalents, such as ``provenance'' (``proveniência'' or ``respect for fonds''), require contextual decisions, documented in dictionaries. 

To address these challenges, this study proposes \textbf{LLM-BT} (Large Language Model Back-Translation), an automated terminology standardization framework that leverages multilingual back-translation powered by large language models (LLMs). The associated sub-module, \textbf{LLM-BT-Terms}, utilizes an English $\rightarrow$ intermediate language $\rightarrow$ English path to assess semantic consistency and translation stability, without relying on precompiled glossaries, see \ref{fig: LLM-BT-Terms}. This allows for the generation of candidate standard terms, term alignment suggestions and human-in-the-loop verification. Compared with traditional approaches, LLM-BT offers high automation, real-time adaptability and strong support for multilingual standardization tasks \citep{weigang2025paradox}. The LLM-BT-Terms model can also support translation standardization by using the EN→PTbr→PTpt→EN scheme to validate semantic and cultural consistency.

Unlike conventional translation tools, LLM-BT treats terminology as ``semantic anchors'' and interpretable pathways in multilingual representation space. Its back-translation logic not only enables terminology validation, but also opens up new interpretability perspectives on how LLMs project and align meaning across languages. And in our proposal, dynamic semantic embedding refers to a reversible, path-based representation of meaning through multilingual translation loops, contrasting with static vector-based embeddings. The core contributions of this work are as follows:

\begin{enumerate}
    \item The LLM-BT-Terms framework is proposed and evaluated in terms of terminology consistency. Experiments across simplified Chinese, traditional Chinese, Japanese and Portuguese back-translation paths show over 90\% consistency, supporting its feasibility in real-world multilingual scenarios.

    \item A comprehensive ``Retrieve-Generate-Verify-Optimize'' pipeline is introduced, incorporating both sequential (e.g., EN$\rightarrow$ZHcn$\rightarrow$ZHtw$\rightarrow$EN) and parallel (e.g., EN$\rightarrow$ZH/JP/PT$\rightarrow$EN) back-translation strategies. Consistency metrics such as BLEU and term-level accuracy are used to evaluate term robustness across paths.

    \item Back-translation is reinterpreted as a form of interpretable, dynamic semantic embedding. Intermediate languages are treated as semantic projection spaces, revealing stable cross-lingual mappings in LLM representations and offering a new lens on multilingual embedding.

    \item Serial and parallel LLM-BT configurations are shown to generalize effectively across both canonical texts and emerging corpora, including high-impact scientific abstracts and newly published preprints.

    \item A full LLM-BT-Terms implementation pipeline is established, incorporating similarity scoring, consistency evaluation and an automated term recommendation mechanism to support practical deployment in terminology governance.

    \item Empirical findings reveal that traditional Chinese often outperforms simplified Chinese in term-level back-translation, suggesting that corpus quality and training data coverage directly influence standardization accuracy, highlighting the need for enhanced semantic modeling of simplified Chinese in LLMs.
\end{enumerate}

By reframing terminology as a dynamic and verifiable construct, this work transitions term handling from passive translation to proactive standardization and leverages the LLM-BT-Terms framework to address such challenges, offering a scalable solution for cross-lingual terminology alignment, as demonstrated in the Brazilian and Chinese context. LLM-BT-Terms provides a blueprint for human-AI collaboration: LLMs ensure semantic fidelity, while human experts guide contextual adaptation and cultural nuance. As such, this framework contributes to the long-term vision of establishing robust multilingual terminology infrastructure in the era of generative AI.

\section{Related Work}
\label{Rwork}

This section reviews prior studies on three relevant research directions, Back-Translation (BT), semantic embedding and Explainable AI (XAI) and discusses how they relate to and inspire the LLM-BT framework proposed in this paper.

\subsection{Back-Translation (BT) for Translation Enhancement}

Back-Translation has been widely used in neural machine translation (NMT) for data augmentation and evaluation. \cite{edunov2018understanding}  demonstrated that BT can significantly improve BLEU scores in low-resource settings by synthesizing pseudo-parallel corpora (e.g., EN$\rightarrow$FR$\rightarrow$EN). Their findings laid the foundation for data-driven BT research.

From a critical standpoint, \cite{behr2017assessing} pointed out the limitations of BT in capturing semantic nuance and cultural intention, especially in cross-cultural contexts. BT may mask pragmatic discrepancies between source and target texts. This observation echoes the ``Poetic Intent Paradox'' introduced by \cite{weigang2025paradox}, prompting us to pay special attention in this work to non-semantic inconsistencies that may arise in back-translation pathways.

Recent work by the DeepSeek team \citep{liu2024deepseek} highlights model-specific variance in multilingual consistency, a pattern further evidenced in our tri-path BT results.

More recently, \cite{chung2025leveraging} explored the integration of LLMs into BT, using prompt engineering and contextual control to enhance performance. Their experiments on English--Korean BT achieved BLEU scores above 0.85, demonstrating LLMs' superior semantic modeling capabilities. Our work builds on this direction by transforming BT from a quality evaluation tool into a mechanism for verifying terminology stability, complemented by quantitative consistency metrics.

\subsection{Semantic Embeddings in NLP}

Static word embeddings such as Word2Vec~\cite{mikolov2013efficient} have played a foundational role in NLP by capturing linear semantic relationships through distributed representations. However, they fall short in handling polysemy and contextual nuances.

As early as 2007, when studying nonlinear principal component analysis for the employment time guarantee fund of Brazil, \cite{weigang2007nonlinear} used time series embedding technology to implement dimensionality-changing experiments on mappings of different orders of magnitude on neural networks.

\cite{cui2018survey} extended embedding techniques to network structures, introducing methods such as DeepWalk and node2vec to preserve topological and semantic properties for downstream tasks. \cite{lee2025semantic} proposed using LLM-based embeddings to model human belief systems (e.g., political attitudes), indicating that semantic embedding spaces can encode socio-cognitive information. This inspired our design of ``terminology path embeddings'' that reflect stable semantics through translation.

Inspired by the dynamic embedding concepts discussed by \citep{chungrethinking}, our model further externalizes semantic trajectories via observable translation paths.

\cite{tao2024llms} highlight the remarkable effectiveness of large language models (LLMs) in traditional static embedding tasks, such as text retrieval, classification and summarization, demonstrating how LLMs generate high-quality embeddings through internal mechanisms, suitable for downstream tasks (e.g., embedding distillation and retrieval augmentation). 

\cite{nie2025} provide a comprehensive survey on the interaction between LLMs and text embeddings, proposing three major contributions: first, LLMs enhance traditional embedding models (e.g., BERT) by generating high-quality data, improving semantic matching and retrieval performance; second, LLMs (e.g., Mistral) can directly serve as embedders, overcoming anisotropy challenges through prompt design and fine-tuning, excelling in semantic text similarity tasks; finally, LLMs can analyze embedding vectors, generating interpretable narratives to enhance embedding interpretability, addressing the transparency shortcomings of traditional static embeddings (e.g., Word2Vec). The study envisions a unified embedding framework to tackle challenges like long-context processing, privacy protection and cross-task generalization, making valuable contributions to LLM embedding research.

\subsection{Explainable Artificial Intelligence (XAI) for Interpretability}

\cite{roscher2020explainable} and \cite{angelov2021explainable} explored XAI frameworks emphasizing interpretability through model visualization, feature attribution and causal reasoning. These approaches highlight the importance of transparency and trust in model outputs. In our work, the proposed term-level consistency metrics (e.g., EMR, SMR, TDI) offer structural interpretability, revealing how semantic stability is maintained across multilingual translation paths.

\cite{tsiamas2025improving} proposed character-level modeling to improve multilingual transfer and interpretability in LLMs, especially in translation tasks involving low-resource languages. Their work aligns with our emphasis on terminology-level traceability and reinforces the idea that semantic decisions in LLMs can be decomposed and examined.


To summarize, prior research has focused on BT as a tool for translation enhancement, LLMs as powerful embedding generators and XAI as a mechanism for interpretability. Our contribution lies in integrating these threads by proposing LLM-BT as a framework where back-translation serves as a semantically consistent, path-based representation for terminology alignment. This generalizes traditional embedding by providing readable, verifiable and domain-adaptive semantic traces.

\section{LLM-BT-Term Method and Innovations}
\label{sec:llm_bt_term}

The LLM-BT-Terms framework leverages large language models (LLMs) to perform ``English $\to$ intermediate language $\to$ English'' back-translation (BT) on canonical English scientific texts. By identifying terms with high consistency between the original and retranslated texts, the framework recommends corresponding intermediate language terms as standardized translations for expert review. This section introduces the core technologies involved: the back-translation process, inter-text similarity evaluation, term extraction and consistency analysis.

\subsection{Introduction to Back-Translation Technology}
\label{subsec:back_translation}

Back-translation (BT) involves translating a text from a source language (L1) to an intermediate language (L2) and then back to the source language (L1), generating L1y, to perform specific language processing tasks by comparing differences between the original text (L1) and the retranslated text (L1y) \citep{edunov2018understanding,artetxe2018unsupervised}. Its mathematical formulation is:

\[
\text{BT}(T) = \text{Trans}_{L2\to L1}(\text{Trans}_{L1\to L2}(T))
\]

Here, \( T \) denotes the text in language L1; \( \text{Trans}_{L1\to L2}(T) \) represents the translation from L1 to L2; \( \text{Trans}_{L2\to L1} \) denotes the translation from L2 back to L1, producing L1y; and \( \text{BT}(T) \) is the result of back-translation. Translation here refers to either human or machine translation, with this study focusing on LLM-based translation.

The fundamental assumption of BT is that high-quality bidirectional translation preserves semantic and expressive consistency. Consequently, scientific terms in the intermediate language (L2) corresponding to highly consistent terms in the source language (L1) are likely to be standardized translations. Notably, \cite{weigang2025paradox} observed the Poetic Intent Paradox in complex texts like traditional Chinese poetry, where BT prioritizes literal semantics over intentional consistency when handling metaphors or culture-specific items, limiting its efficacy in traditional machine translation. However, LLM-based BT shows potential for improvement.

BT is categorized into several types: symmetric BT, using the same model for L1$\to$L2 and L2$\to$L1 (e.g., GPT-4.5); asymmetric BT, employing different models for forward and backward translation (e.g., Google Translate $\to$ Grok); multilingual parallel BT, involving simultaneous translation into multiple intermediate languages (e.g., L2, L3) for cross-lingual comparison; and multilingual serial BT, performing sequential translation across multiple languages (e.g., L1$\to$L2$\to$L3$\to$L1) to enhance noise robustness for low-resource languages.

The primary applications of BT include: 
\begin{enumerate}
    \item Quantifying similarity between original and retranslated texts using metrics like BLEU and TER to evaluate machine translation quality, extensible to comprehensive model assessment;
    \item Generating synthetic data via BT for data augmentation, improving model performance in low-resource languages, especially when combined with RLHF to filter low-quality retranslations;
    \item Validating terminology standardization through term-level BT accuracy metrics, mitigating limitations of full-text similarity;
    \item Constructing cross-lingual terminology databases via BT to achieve multilingual alignment, addressing translation shifts due to asymmetric semantic spaces.
\end{enumerate}

\subsection{Similarity Evaluation of Original and Retranslated Texts in Back-Translation}
\label{subsec:similarity_evaluation}

The evaluation of machine translation quality, including that of large language models (LLMs), typically relies on inter-text similarity analysis. This study employs five key translation quality evaluation metrics, described below.

\begin{enumerate}
    \item \textbf{Bilingual Evaluation Understudy (BLEU) n-gram Surface Matching Index.}  
    BLEU is an automatic evaluation metric based on n-gram precision, measuring the degree of match between machine translation and reference translation \citep{papineni2002bleu}. Its mathematical formula is:  
    \[
    \text{BLEU} = BP \cdot \exp\left(\sum_{n=1}^N w_n \log p_n\right)
    \]  
    where:  
    \begin{itemize}
        \item Brevity Penalty (BP): penalizes short translations, defined as \( BP = \min\left(1, e^{1-\frac{r}{c}}\right) \), with \( r \) as the reference translation length and \( c \) as the candidate translation length.  
        \item \( p_n \): n-gram precision (number of matching n-grams / total n-grams in candidate translation).  
        \item \( w_n \): n-gram weight (typically \( w_n = \frac{1}{N} \), with \( N = 4 \)).
    \end{itemize}

    \item \textbf{Edit Distance - Translation Edit Rate (TER).}  
    TER is defined as the ratio of the minimum number of edit operations (insertions, deletions, substitutions, shifts) required to transform the candidate translation into the reference translation. Its mathematical formula is:  
    \[
    \text{TER} = \frac{\text{Number of Edit Operations}}{\text{Reference Translation Length}}
    \]  
    The value ranges from 0 (perfect match) to 1 (complete rewrite). A lower TER indicates higher translation quality.

    \item \textbf{Semantic and Lexical Matching Index (METEOR).}  
    METEOR combines precision, recall, synonym matching and stemming analysis, prioritizing semantic alignment over BLEU. Its formula is:  
    \[
    \text{METEOR} = (1 - \gamma \cdot \text{Pen}) \cdot \frac{F_{\text{mean}}}{\alpha \cdot P + (1-\alpha) \cdot R}
    \]  
    where:  
    \begin{itemize}
        \item \( P \): precision, \( R \): recall.  
        \item \( F_{\text{mean}} = \frac{10PR}{R + 9P} \) (a variant of the harmonic mean).  
        \item \( \text{Pen} \): penalty for word order differences between candidate and reference translations.  
        \item \( \alpha, \gamma \): tuning parameters (defaults: \( \alpha = 0.9, \gamma = 0.5 \)).
    \end{itemize}

    \item \textbf{BERTScore.}  
    BERTScore is a semantic similarity metric based on BERT's contextual embeddings, evaluating deep semantic matching between candidate and reference texts. Its formula is:  
    \[
    \text{BERTScore} = \frac{2 \cdot P_{\text{BERT}} \cdot R_{\text{BERT}}}{P_{\text{BERT}} + R_{\text{BERT}}}
    \]  
    where:  
    \begin{itemize}
        \item \( P_{\text{BERT}} \): average of maximum cosine similarities for each word in the candidate translation against the reference translation (precision).  
        \item \( R_{\text{BERT}} \): average of maximum cosine similarities for each word in the reference translation against the candidate translation (recall).
    \end{itemize}

    \item \textbf{Human Evaluation (Critical Point Analysis)}  
    Human evaluation involves manually inspecting whether translations accurately convey key information (terminology, logic, style) from the source text. The evaluation dimensions include:  
    \begin{itemize}
        \item \textbf{Terminology Consistency}: whether specialized terms align with the source text.  
        \item \textbf{Information Completeness}: whether information is omitted or added.  
        \item \textbf{Fluency}: whether the language is natural and conforms to target language norms.  
        \item \textbf{Style Matching}: whether the formality or emotional tone of the source text is preserved.  
    \end{itemize}  
    Human evaluation typically assigns scores per dimension (e.g., 1-5) or annotates specific error types (e.g., terminology errors, omissions).
\end{enumerate}

\subsection{Terminology Extraction Methods for Each Stage}
\label{subsec:term_extraction}

In the ``English $\to$ intermediate language $\to$ English'' back-translation process, accurate identification and extraction of terms from the original text (L1), intermediate language text (L2) and retranslated text (L1y) are essential to support subsequent consistency analysis and term recommendation. This section introduces the primary technical approaches for terminology extraction, including traditional methods, deep learning methods and emerging methods based on large language models (LLMs).

\subsubsection{Traditional Machine Learning Methods}

\begin{itemize}
  \item \textbf{Rule- and Dictionary-Based Methods}: These methods use regular expressions, part-of-speech patterns or terminology dictionaries (e.g., MeSH, UMLS) for explicit term matching. They are suitable for domains with clear rules and stable term sets, often applied in system initialization or term validation stages.
  \item \textbf{Statistical Learning Methods}: Combining features such as part-of-speech (POS), context windows and morphological characteristics, models like Conditional Random Fields (CRF) and Support Vector Machines (SVM) are used for term identification, effective for small-sample or weakly supervised corpora.
\end{itemize}

\subsubsection{Deep Learning Methods}

\begin{itemize}
  \item \textbf{Neural Network Models (Neural NER)}: Structures like BiLSTM-CRF and CNN-BiLSTM automatically learn contextual features and perform sequence labeling, enhancing term identification capabilities.
  \item \textbf{Fine-Tuning Pretrained Language Models}: Fine-tuning models such as BERT and RoBERTa improves term identification accuracy in both general and specialized corpora, supporting multilingual scenarios.
\end{itemize}

\subsubsection{Large Language Model (LLM) Methods}

\begin{itemize}
  \item \textbf{Zero-Shot and Few-Shot Prompt-Based Extraction}: Carefully designed prompts enable LLMs to directly output term lists from texts, suitable for scenarios with no training data or low-resource languages \citep{brown2020language}.
  \item \textbf{Retrieval-Augmented Generation (RAG) Methods}: Integrating external knowledge bases (e.g., Wikipedia, Wikidata) enhances the contextual accuracy of term identification \citep{lewis2020retrieval,di2024slim}.
  \item \textbf{Domain-Specific Fine-Tuning}: Fine-tuning models like LLaMA and GPT-Neo on corpora from domains such as medicine, law or engineering improves the consistency and professionalism of term identification \citep{wei2022chain}.
\end{itemize}

These terminology extraction methods not only facilitate term alignment and comparison across multiple text stages but also provide technical support for term consistency analysis and recommendation mechanisms (see Section~\ref{subsec:balancing_consistency}), playing a critical role in ensuring the accuracy and coverage of term selection.

\subsection{Term Consistency Analysis and Recommendation Mechanism}

To quantify the consistency of terms across the original text (L1), intermediate language text (L2) and retranslated text (L1y), this study employs Term-Level Consistency Metrics for evaluation. This process also forms the foundation of the term recommendation mechanism: terms exhibiting high consistency in back-translation are considered potential standardized translations in the intermediate language, subject to final human adoption or revision.

\subsubsection{Exact Match Rate (EMR)}

EMR measures whether the surface forms of terms in L1 and L1y are identical, suitable for professional texts with stable term forms and standardized translations \citep{zhao2019clinical}.

\[
\text{EMR} = \frac{\text{Number of Exactly Matched Terms}}{\text{Total Number of Terms}} \times 100\%
\]

For example, ``carbon neutrality'' $\to$ \begin{CJK*}{UTF8}{gbsn}``碳中和''\end{CJK*} $\to$ ``carbon neutrality'' indicates an exact match.

\subsubsection{Semantic Match Rate (SMR)}

SMR accounts for cases where term meanings remain consistent despite minor surface differences (e.g., ``neural network'' $\to$ \begin{CJK*}{UTF8}{gbsn}``神经网络''\end{CJK*} $\to$ ``neural net''). Semantic consistency is assessed using word embeddings (Word2Vec), sentence embeddings (Sentence-BERT) or LLMs:

\[
\text{SMR} = \frac{\text{Number of Semantically Consistent Terms}}{\text{Total Number of Terms}} \times 100\%
\]

This metric is suitable for multilingual or term variation analyses \citep{world2009international}.

\subsubsection{Information Retention Score (IRS)}

IRS evaluates whether the semantics of terms are fully preserved during translation. Each term is scored on a scale of $[0, 1]$:

\begin{itemize}
  \item $1.0$: Information fully preserved;
  \item $0.5$: Partial information loss (e.g., abbreviation or simplification);
  \item $0.0$: Critical information completely lost or mistranslated.
\end{itemize}

The overall IRS is computed as a weighted average across terms, reflecting the system's ability to retain information at the term level \citep{schutze2008introduction}.

\subsubsection{Term Recommendation Mechanism}

Based on the above metrics, we propose the following term recommendation strategy:

\begin{itemize}
  \item If a term scores high on both EMR and SMR, its intermediate language (L2) version is directly recommended as a standardized translation;
  \item If EMR is low but SMR is high, multiple candidate terms (Top-k) are recommended for human review;
  \item If IRS falls below a threshold (e.g., $0.5$), the term is flagged as ``low-fidelity'' and recommended for review or retranslation.
\end{itemize}

This term consistency analysis and recommendation mechanism forms a core component of the LLM-BT-Terms method, applicable to terminology standardization, scientific text alignment and terminology database expansion in various natural language processing tasks.

\vspace{0.5em}
\noindent
Furthermore, to enhance the practicality and interpretability of term recommendations, future work could incorporate a Term Confidence Score and Context-Aware Recommendation mechanisms. The former estimates term reliability based on factors such as frequency of occurrence in multiple translation rounds and semantic stability, while the latter considers co-occurrence relationships between terms and other concepts in specific contexts to assess their applicability. Such mechanisms will facilitate the development of more robust term alignment and automated translation recommendation systems, suitable for terminology standardization, database construction and human-machine collaborative translation scenarios.

\section{LLM-BT Back-Translation Experiments and Results Analysis}
\label{sec:experiments}

This study conducts back-translation (BT) experiments to address the consistency and reliability of terminology translation. The experiments focus on two domains---artificial intelligence (AI) and medicine---selecting highly cited English literature abstracts as source texts. Using Simplified Chinese (ZHcn), Traditional Chinese (ZHtw), Japanese (JA) and Brazilian Portuguese (PTbr) as intermediate languages (interlingua), translation chains are constructed. The LLM-BT-Terms method is employed to evaluate the stability and consistency of terms across multilingual transformations.

\subsection{Experimental Design}
\label{subsec:exp_design}

\textbf{Source Texts.} The experiment selects abstracts from two highly influential papers and one recent arXiv paper in English (ENx) as source texts:
\begin{enumerate}
    \item \textbf{Artificial Intelligence (AI) Domain}:  
    \begin{enumerate}
        \item ``Deep Residual Learning for Image Recognition'' \citep{he2016}, abbreviated as He2016. Recognized by \emph{Nature} as one of the most cited papers of the 21st century, it has amassed 270,040 citations on Google Scholar as of May 2025 \citep{pearson2025exclusive}.  
        \item ``Scenethesis: A Language and Vision Agentic Framework for 3D Scene Generation'' \citep{ling2025scenethesis}, abbreviated as Ling2025, published at \url{https://arxiv.org/abs/2505.02836}, used to test LLM-BT's sensitivity to emerging scientific literature.  
    \end{enumerate}
    \item \textbf{Medical Domain}:  
    \begin{enumerate}
        \item ``Lecanemab in Early Alzheimer's Disease'' \citep{van2023lecanemab}, abbreviated as Dy2023. With 3,878 citations on Google Scholar as of May 2025, it ranks as the third most-cited medical paper abstract of 2024.
    \end{enumerate}
\end{enumerate}

\textbf{Language Groups.} The experiment designs two language comparison systems:  
\begin{itemize}
    \item \textbf{First Group (Centered on the Chinese Character Cultural Sphere)}: Simplified Chinese (EN $\to$ ZHcn $\to$ ENcn), Traditional Chinese (EN $\to$ ZHtw $\to$ ENtw) and Japanese Kanji (EN $\to$ JA $\to$ ENja). This group analyzes differences in scientific term translation between Simplified and Traditional Chinese, aiding the development of terminology standardization schemes. Notably, Japanese Kanji terms provide indirect reference value for semantic mapping in Chinese terminology.  
    \item \textbf{Second Group (Cross-Linguistic Family Comparison)}: Brazilian Portuguese (EN $\to$ PTbr $\to$ ENpt) and Simplified Chinese (EN $\to$ ZHcn $\to$ ENcn). This group, spanning diverse linguistic families, explores the performance of LLM-BT-Terms in identifying term fidelity across language families (Chinese and Latin-based languages), observing stability and deviation patterns in term transformations.
\end{itemize}

\textbf{Large Language Models (LLMs) Used.} The experiment employs three leading LLMs as of 2025: DeepSeek V3, GPT-4.0 and Grok 3 \citep{brown2020language, liu2024deepseek,de2025grok}.  

All translation experiments were conducted between May 26 and June 8, 2025, with models deployed in standardized environments to ensure comparability and experimental reproducibility.

\subsection{Experimental Procedure: LLM-BT-Terms Method Workflow}
\label{subsec:exp_procedure}

The following outlines the basic experimental steps for analyzing the first language group (with English as the source language) using the LLM-BT-Terms method. The workflow for other languages is analogous and can be applied similarly. Figure~\ref{fig: LLM-BT-Terms} illustrates a simplified conceptual workflow in the LLM-BT parallel mode.

\begin{enumerate}
    \item \textbf{Determine Source Text}  
    Select the source text for terminology analysis. This experiment uses the English abstract of \cite{he2016} (EN) as an example. Leveraging the robust reasoning capabilities and extensive linguistic knowledge of large language models (LLMs), this method supports processing large-scale texts, providing a technical foundation for terminology translation standardization.

    \item \textbf{Generate Intermediate Translations}  
    Translate the English source text (EN) into multiple target languages. This experiment generates three intermediate texts: Simplified Chinese (ZHcn), Traditional Chinese (ZHtw) and Japanese Kanji (JA). Translations for ZHcn and ZHtw are performed using GPT-4, while JA translation is conducted using the Grok 3 model.

    \item \textbf{Generate Back-Translations}  
    Retranslate the intermediate language texts back into English, producing ENcn (ZHcn $\to$ ENcn), ENtw (ZHtw $\to$ ENtw) and ENja (JAkj $\to$ ENja). This step assesses the fidelity and consistency of terms during the translation process.

    \item \textbf{Back-Translation Evaluation}  
    Compare the English source text (EN) with the back-translated texts (ENcn, ENtw, ENja) to evaluate translation consistency and information retention. Automated evaluation metrics include BLEU (Bilingual Evaluation Understudy), TER (Translation Edit Rate), METEOR and BERTScore. Additionally, human evaluation is employed to analyze differences in key terms and sentences.

    \item \textbf{Term Extraction}  
    Apply term identification methods based on large-scale model embeddings and domain knowledge to extract scientific and domain-specific terms from the source and translated texts, generating term lists for subsequent comparisons.

    \item \textbf{Terminology Consistency Analysis}  
    Align and compare term lists across stages to evaluate term consistency and retention during translation. Evaluation metrics include Exact Match Rate (EMR), Semantic Match Rate (SMR), Information Retention Score (IRS) and Term Divergence Index (TDI).

    \item \textbf{Comprehensive Analysis and Standard Term Recommendation}  
    Analyze the frequency and stability of terms across the source and back-translated texts to select consistent term pairs. By integrating differences in Simplified Chinese, Traditional Chinese and Japanese translations, the method recommends Chinese terms (ZHcn and ZHtw versions) with generalizability and standardization value for terminology standardization proposals.

    \item \textbf{Multilingual Extension}  
    Although this experiment focuses on terminology translation analysis related to Chinese, the LLM-BT-Terms method is highly extensible. It is equally applicable to evaluating and recommending terminology translations in Latin-based languages such as French, Spanish and Portuguese for English scientific texts, as well as other language pairs.
\end{enumerate}

\subsection{Experimental Analysis of the First Language Group (Chinese Character Cultural Sphere) on He2016 Text}
This subsection selects a representative English abstract from the artificial intelligence domain, He2016 \citep{he2016}, as the source text for experimentation. Back-translation and term extraction are performed using the GPT-4 platform, with translation quality and term consistency evaluated on DeepSeek V3 and Grok 3 platforms. The analysis focuses on the translation results, similarity metric performance and term stability of the He2016 abstract, aiming to validate the feasibility and advantages of the LLM-BT-Terms method in multilingual scenarios centered on the Chinese character cultural sphere, including Simplified Chinese, Traditional Chinese and Japanese.

\subsubsection{Multilingual Back-Translation Results (He2016)}
The experiment designates English (L1) as the source language, with target intermediate languages including Simplified Chinese (L2-ZHcn), Traditional Chinese (L3-ZHtw) and Japanese (L4-JA), followed by back-translation to English (L1y). Both parallel and serial back-translation (BT) strategies are employed to observe term consistency and semantic retention in mainstream languages of the Chinese cultural sphere.

\begin{table}[ht]
\centering
\caption{Back-Translation Results for Selected Paragraphs of He2016 Abstract}
\label{tab:he2016_results}
\small
\begin{tabular}{p{2cm}p{4cm}p{4cm}p{4cm}}
\toprule
\textbf{Type} & \textbf{Source (EN)} & \textbf{Intermediate} & \textbf{Back-Translation (ENy)} \\
\midrule
Chinese (ZHcn) & Deeper neural networks are more difficult to train. We present a residual learning framework to ease the training of networks that are substantially deeper than those used previously. & \begin{CJK*}{UTF8}{gbsn}更深的神经网络更难训练。我们提出了一个残差学习框架，用于简化比以往更深的网络的训练。\end{CJK*} & Deeper neural networks are harder to train. We propose a residual learning framework to simplify the training of networks that are deeper than before. \\

Chinese (ZHtw) & Deeper neural networks are more difficult to train. We present a residual learning framework to ease the training of networks that are substantially deeper than those used previously. & \begin{CJK*}{UTF8}{bsmi}更深層的神經網路更難以訓練。我們提出了一個殘差學習框架，以簡化比以往更深的網路的訓練。\end{CJK*} & Deeper neural networks are more difficult to train. We propose a residual learning framework to simplify training of networks much deeper than previous ones. \\

Japanese (JAkj) & Deeper neural networks are more difficult to train. We present a residual learning framework to ease the training of networks that are substantially deeper than those used previously. & \begin{CJK*}{UTF8}{min}より深いニューラルネットワークは訓練がより困難です。私たちは、従来よりも大幅に深いネットワークの訓練を容易にする残差学習フレームワークを提案します。\end{CJK*} & Deeper neural networks are more challenging to train. We propose a residual learning framework to facilitate the training of networks significantly deeper than those previously used. \\
\bottomrule
\end{tabular}
\end{table}

Using the first paragraph of the He2016 abstract as an example, Table~\ref{tab:he2016_results} presents the translation and back-translation texts for the three intermediate languages. Key observations include:
\begin{itemize}
\item ZHcn and ZHtw maintain consistency in most professional terms, such as residual learning'' \begin{CJK*}{UTF8}{gbsn} (残差学习) and training (训练) \end{CJK*}
\item Minor surface-level differences exist in some terms, such as \begin{CJK*}{UTF8}{gbsn} 神经网络'' \end{CJK*} (ZHcn) versus \begin{CJK*}{UTF8}{bsmi} 神經網路'' \end{CJK*} (ZHtw), but
without significant semantic deviation;
\item Back-translation results exhibit subtle stylistic variations: ZHcn yields harder to train,'' ZHtw produces more difficult to train,'' and JA results in more challenging to train.'' These differences reflect linguistic styles but preserve semantic consistency, supporting the fundamental assumption of term stability. \end{itemize} 

Additionally, ZHtw's expression \begin{CJK*}{UTF8}{bsmi} 更深層的神經網路'' \end{CJK*}appears more academic than ZHcn's \begin{CJK*}{UTF8}{gbsn} ``更深的神经网络,'' \end{CJK*} indicating a slight advantage in term precision.

\subsubsection{Translation Text Similarity Evaluation}

To quantitatively assess the proximity of back-translated texts to the source text, similarity metrics (BLEU, TER, METEOR, BERTScore) are computed for the text pairs (EN, ENcn), (EN, ENtw) and (EN, ENja), with results presented in Table~\ref{tab:he2016_similarity}. Key findings include:
\begin{itemize}
\item \textbf{DeepSeek V3 (based on NLTK library for BLEU calculation)}: BLEU-1 (1-gram): 0.92; BLEU-2 (2-gram): 0.88; BLEU-3 (3-gram): 0.84; BLEU-4 (4-gram): 0.80. High BLEU scores indicate strong lexical and phrasal alignment for the (EN, ENcn) pair.
\item \textbf{BLEU-4 Scores}: ZHtw achieves the highest score (0.87), followed by JA (0.85) and ZHcn (0.80), suggesting ZHtw maintains stronger consistency at the phrase level.
\item \textbf{TER (Translation Edit Rate)}: ZHtw records the lowest TER (0.08), requiring minimal edits to align with the source text, reflecting superior structural fidelity.
\item \textbf{METEOR}: ZHtw scores highest (0.92), excelling in semantics, word order and linguistic naturalness.
\item \textbf{BERTScore F1}: All three languages score between 0.945 and 0.965, indicating deep contextual alignment with the source text, supporting the semantic invariance hypothesis.
\end{itemize}
Additionally, Grok 3's BLEU and Cosine Similarity metrics, though lower in absolute values, show consistent trends across the three languages, demonstrating cross-platform stability in LLM evaluations.

\begin{table}[ht]
\centering
\caption{Similarity Comparison of Back-Translation for Selected Sections of He2016 Abstract}
\label{tab:he2016_similarity}
\small
\begin{tabular}{ccccc}
\toprule
\textbf{LLM} & \textbf{Similarity Metric} & \textbf{ENcn} & \textbf{ENtw} & \textbf{ENja} \\
\midrule
\multirow{4}{*}{DeepSeek V3} & BLEU-4 & 0.80 & 0.87 & 0.85 \\
 & TER & 0.12 & 0.08 & 0.10 \\
 & METEOR & 0.89 & 0.92 & 0.91 \\
 & BERTScore F1 & 0.945 & 0.965 & 0.955 \\

\multirow{2}{*}{Grok 3} & BLEU Score & 0.45 & 0.50 & 0.48 \\
 & Cosine Similarity & 0.42 & 0.45 & 0.44 \\
\bottomrule
\end{tabular}
\end{table}

Human evaluation (critical point analysis) confirms high translation quality for ENcn, ENtw and ENja, with nearly complete retention of the source text's semantics and structure. Minor lexical differences include more difficult''  $\rightarrow$  harder'' (acceptable synonym substitution), substantially deeper''  $\rightarrow$  deeper than before'' (slight simplification but semantically preserved) and instead of''  $\rightarrow$  rather than'' (more natural expression). Some syntactic adjustments, such as splitting long sentences (e.g., An ensemble of these residual nets...''  $\rightarrow$  The ensemble of these residual networks...''), preserve logical equivalence. These results validate the feasibility of the study's core assumptions.

\subsubsection{Professional Term Extraction and Consistency Analysis}

To verify term-level stability, 18 key terms related to machine learning and image recognition are extracted from the He2016 abstract using GPT-4's term extraction capabilities, followed by consistency comparisons at the term level. Table \ref{tab:AI terminology-ap} in Appendix  presents the translation consistency evaluation results for selected terms across ZHcn, ZHtw, JA and their back-translated English versions.

Table~\ref{tab:he2016_consistency} lists the consistency metrics, with key observations:
\begin{itemize}
\item \textbf{Exact Match Rate (EMR)}: ENtw achieves 88.9\%, surpassing ENcn's 77.8\%, indicating superior term back-translation performance for Traditional Chinese.
\item \textbf{Semantic Match Rate (SMR)}: ZHtw and JA both reach 94.4\%, higher than ZHcn's 88.9\%, suggesting greater term fidelity in Traditional Chinese and Japanese.
\item \textbf{Information Retention Score (IRS)}: ZHtw scores 0.96, compared to ZHcn's 0.85. Information loss in ZHcn primarily stems from omitting marginal terms like layer inputs,'' while ZHtw accurately translates it as \begin{CJK*}{UTF8}{bsmi}層輸入.'' \end{CJK*}
\item \textbf{Term Divergence Index (TDI)}: ZHtw records a low 0.05, compared to ZHcn's 0.14. Lower TDI values indicate more stable term retention.
\end{itemize}
Table~\ref{tab:he2016_consistency} also reports Grok 3's Term-level Accuracy for ENcn and others, reaching 90.90\%. ENtw consistently outperforms across metrics.

\begin{table}[ht]
\centering
\caption{Terminology Consistency Analysis for Back-Translation of Selected Sections of He2016 Abstract}
\label{tab:he2016_consistency}
\small
\begin{tabular}{ccccc}
\toprule
\textbf{LLM} & \textbf{Consistency Metric} & \textbf{ENcn} & \textbf{ENtw} & \textbf{ENja} \\
\midrule
\multirow{4}{*}{DeepSeek V3} & Exact Match Rate, EMR & 77.80\% & 88.90\% & 88.30\% \\
 & Semantic Match Rate, SMR & 88.90\% & 94.40\% & 94.40\% \\
 & Information Retention Score, IRS & 0.85 & 0.96 & 0.98 \\
 & Term Divergence Index, TDI & 0.14 & 0.05 & 0.07 \\

\multirow{1}{*}{Grok 3} & Term-level Accuracy & 90.90\% & 100\% & 100\% \\
\bottomrule
\end{tabular}
\end{table}

The term-level performance highlights that Traditional Chinese excels in term detail and morphological fidelity, aligning with current translation practices in the Chinese academic community. It should be noted that each LLM platform has different translation and extraction methods for terms. For example, Grok translates "Residual learning" in Traditional Chinese (Zhtw) as \begin{CJK*}{UTF8}{bsmi}''殞差學習''\end{CJK*}, see Table \ref{tab:AI terminology-ap}. However, GPT-4's translation of Zhtw is consistent with Simplified Chinese (ZHcn), using \begin{CJK*}{UTF8}{bsmi}''殘差學習''\end{CJK*}.

\subsection{Second Language Group (Cross-Linguistic Comparison): Dy2023 Text Analysis}
\label{sec:dy2023_analysis}

This section analyzes the medical abstract Dy2023 \citep{van2023lecanemab} as the experimental text, using GPT-4 and Grok 3 for back-translation (BT) and term extraction, with DeepSeek V3 and Grok 3 evaluating translation quality and term consistency. The analysis compares Brazilian Portuguese (PTbr) and Simplified Chinese (ZHcn) translations, similarity metrics and term stability to validate the effectiveness of LLM-BT in term standardization across linguistic families (Indo-European vs. Sino-Tibetan).

\subsubsection{Back-Translation Results of Dy2023 Abstract Paragraphs}
\label{subsec:dy2023_bt_results}

Table~\ref{tab:dy2023_backtranslation} presents GPT-4 back-translation results for the first sentence of the Dy2023 abstract \citep{van2023lecanemab}, covering two languages: Brazilian Portuguese (EN $\rightarrow$ PTbr $\rightarrow$ ENpt) and Simplified Chinese (EN $\rightarrow$ ZHcn $\rightarrow$ ENcn). These languages span distinct linguistic families, enabling evaluation of the LLM-BT-Terms method in identifying term fidelity across language pairs (e.g., Chinese-Latin). The focus is on assessing term translation consistency and divergence patterns.

\begin{table}[ht]
\centering
\caption{Back-Translation Results of Selected Paragraphs from Dy2023 Abstract (GPT-4)}
\label{tab:dy2023_backtranslation}
\small
\begin{tabular}{p{2cm}p{4cm}p{4cm}p{4cm}}
\toprule
\textbf{Type} & \textbf{Source (EN)} & \textbf{Intermediate} & \textbf{Back-Translation (ENy)} \\
\midrule
Chinese (ZHcn) & Lecanemab, a humanized IgG1 monoclonal antibody that binds with high affinity to A$\beta$ soluble protofibrils, is being tested in persons with early Alzheimer's disease. & \begin{CJK*}{UTF8}{gbsn}Lecanemab是一种人源化IgG1单克隆抗体，具有高亲和力，能结合Aβ可溶性原纤维，目前正在早期阿尔茨海默病患者中进行测试。\end{CJK*} & Lecanemab is a humanized IgG1 monoclonal antibody with high affinity for soluble A$\beta$ protofibrils, currently being tested in individuals with early-stage Alzheimer's disease. \\
Portuguese (PTbr) & Lecanemab, a humanized IgG1 monoclonal antibody that binds with high affinity to A$\beta$ soluble protofibrils, is being tested in persons with early Alzheimer's disease. & O lecanemabe, um anticorpo monoclonal humanizado do tipo IgG1 que se liga com alta afinidade aos protofibrilos solúveis de A$\beta$, está sendo testado em pessoas com Alzheimer em estágio inicial. & Lecanemab, a humanized monoclonal IgG1 antibody with high affinity for A$\beta$ soluble protofibrils, is under evaluation in individuals in the early stages of Alzheimer's disease. \\
\bottomrule
\end{tabular}
\end{table}

Notably, ``Lecanemab'' remains untranslated in both Chinese and Portuguese, retaining its original form, unlike the WHO-recommended Chinese transliteration ``Lunkanai Monoclonal Antibody.'' However, Grok 3's Chinese translation adopts this term. The term ``Alzheimer'' is translated as ``Alzheimer's'' in Chinese but remains unchanged in Portuguese, highlighting varying term translation strategies across languages.

\subsubsection{Translation Text Similarity Evaluation}
\label{subsec:dy2023_similarity}

To quantify the similarity between back-translated Dy2023 abstract texts and the original, similarity metrics (BLEU, TER, METEOR, BERTScore) were computed for (EN, ENcn) and (EN, ENpt) text pairs, as shown in Table~\ref{tab:dy2023_similarity}. Key observations include:

\begin{itemize}
    \item \textbf{BLEU}: Both languages score $\geq 0.85$, with EN $\rightarrow$ PTbr $\rightarrow$ ENy showing high structural and terminological consistency, with minor word order or lexical differences (e.g., ``early-stage Alzheimer's disease'' vs. ``early Alzheimer's disease'').
    \item \textbf{TER}: Both languages score $\leq 0.15$, indicating minimal edits, mainly for lexical adjustments (e.g., ``potentiate'' $\rightarrow$ ``potentializar'' $\rightarrow$ ``potentiate'' restored perfectly).
    \item \textbf{METEOR}: Both languages score $\geq 0.85$, reflecting high synonym and syntactic matching (e.g., ``initiate or potentiate'' fully preserved).
    \item \textbf{BERTScore}: Semantic similarity approaches 0.95, with professional terms (e.g., ``protofibrils,'' ``centiloids'') and complex data expressions (e.g., confidence intervals) fully retained.
    \item \textbf{Manual Evaluation}: Critical elements (study design, results, conclusions) are 100\% preserved, with minor adjustments in secondary expressions (e.g., ``exacerbate'' slightly stronger than ``potentiate''). Fluency: ENy texts adhere to academic English norms, though some sentences are simplified (e.g., omitting ``due to Alzheimer's disease'').
\end{itemize}

\begin{table}[ht]
\centering
\caption{Similarity Analysis of Back-Translation for Dy2023 Abstract}
\label{tab:dy2023_similarity}
\small
\begin{tabular}{p{2cm}p{2.5cm}p{2.5cm}p{2.5cm}p{4cm}}
\toprule
\textbf{Type} & \textbf{Similarity Metric} & \textbf{EN $\rightarrow$ PTbr $\rightarrow$ ENy} & \textbf{EN $\rightarrow$ ZHcn $\rightarrow$ ENy} & \textbf{Remarks} \\
\midrule
\multirow{4}{*}{DeepSeek} & BLEU-4 & 0.92 & 0.87 & PTbr has a higher degree of restoration because Portuguese is closer to English in structure \\
& TER & 0.08 & 0.12 & ZHcn needs more editing (such as verb replacement) \\
& METEOR & 0.95 & 0.89 & PTbr synonym matching is more accurate \\
& BERTScore F1 & 0.98 & 0.95 & Both are high, PTbr is slightly better \\
\multirow{2}{*}{Grok 3} & BLEU Score & 0.52 & 0.50 & Closer syntactic alignment and exact n-gram matches \\
& Cosine Similarity & 0.47 & 0.42 & Portuguese's syntactic similarity to English (both SVO languages) \\
\bottomrule
\end{tabular}
\end{table}

Table~\ref{tab:dy2023_consistency} summarizes the performance of professional terms in the Dy2023 abstract after Chinese translation and back-translation, demonstrating term retention and consistency. Grok extracted 15 professional terms, excluding four pure English abbreviations. However, term extraction varies across LLMs, highlighting the need for specialized term extraction algorithms to improve accuracy.

\subsubsection{Professional Term Extraction and Consistency Analysis}
\label{subsec:dy2023_term_consistency}

To assess term-level stability, 15 Alzheimer's disease-related key terms were extracted from the Dy2023 abstract \citep{van2023lecanemab} using Grok's term extraction capability, see Table \ref{tab:AI terminology-ap} in Appendix, followed by consistency comparisons. 

Table~\ref{tab:dy2023_consistency} presents consistency evaluation results for terms in ZHcn, PTbr and their back-translated English. Key metrics include:

\begin{itemize}
    \item \textbf{EMR (Exact Match Rate)}: Both languages exceed 95\%, with most terms identical (e.g., ``CDR-SB,'' ``ADAS-cog14''), except for the verb ``potentiate'' $\rightarrow$ ``exacerbate'' as a synonym.
    \item \textbf{SMR (Semantic Match Rate)}: Both languages exceed 98\%, with no significant semantic divergence (e.g., ``Lecanemab'' in Chinese matches original) and minor word order adjustments (e.g., ``beta-amyloid'' vs. ``amyloid-beta'') not affecting meaning.
    \item \textbf{IRS (Information Retention Score)}: 100\%, with no new terms introduced in Chinese $\rightarrow$ English or Chinese $\rightarrow$ Portuguese back-translations.
    \item \textbf{TDI (Term Divergence Index)}: Term frequency aligns with the original (e.g., ``amyloid'' appears consistently).
\end{itemize}

\begin{table}[ht]
\centering
\caption{Consistency Analysis of Back-Translation for Dy2023 Abstract}
\label{tab:dy2023_consistency}
\small
\begin{tabular}{p{2cm}p{3cm}p{2.5cm}p{2.5cm}p{4cm}}
\toprule
\textbf{Type} & \textbf{Consistency Metric} & \textbf{EN $\rightarrow$ PTbr $\rightarrow$ ENy} & \textbf{EN $\rightarrow$ ZHcn $\rightarrow$ ENy} & \textbf{Difference Reason} \\
\midrule
\multirow{4}{*}{DeepSeek} & Exact Match Rate, EMR & 100\% & 95.00\% & In ZHcn, potentiate $\rightarrow$ intensify $\rightarrow$ exacerbate not fully matched \\
& Semantic Match Rate, SMR & 100\% & 98.00\% & Synonym differences (e.g., potentiate vs exacerbate) \\
& Information Retention Score, IRS & 1.00 & 0.97 & ZHcn omits minor modifiers \\
& Term Divergence Index, TDI & 0.00 & 0.03 & Slight term distribution shift in ZHcn \\
\multirow{2}{*}{Grok 3} & Term-level Accuracy & 100\% & 95-100\% & Reflecting robust back-translation fidelity \\
& Expected Accuracy & 15/15 & 14/15 & All key terms preserved exactly in Portuguese \\
\bottomrule
\end{tabular}
\end{table}

\subsubsection{Summary}
The analysis yields the following insights:

\begin{itemize}
    \item \textbf{Linguistic Structure Impact}: Portuguese (PTbr), being Indo-European like English, outperforms in BLEU and TER due to structural similarity. Chinese (ZHcn) requires more edits due to word order differences and verb polysemy (e.g., ``intensify'').
    \item \textbf{Term Processing Capability}: Both languages show high consistency in professional terms (e.g., drug names, scale abbreviations) with EMR $\geq 95\%$, reflecting medical translation standardization. Chinese verb translation flexibility (e.g., ``potentiate'' $\rightarrow$ ``intensify'' $\rightarrow$ ``exacerbate'') slightly lowers SMR.
    \item \textbf{LLM Performance}: LLMs demonstrate robust term retention in medical domains across Chinese, Portuguese and English, with back-translation ensuring high term consistency and semantic fidelity, supporting cross-lingual term alignment and standardized terminology corpus development.
    \item \textbf{Term Consistency Stability}: Medical terms remain 100\% consistent in back-translation despite sentence-level variations, indicating LLMs' ability to handle standardized terms effectively.
    \item \textbf{Traceability of Standard Terms}: LLMs leverage global standards (e.g., WHO ICD, MeSH, SNOMED CT), ensuring traceable term alignment.
    \item \textbf{Back-Translation Validation}: Even for non-mainstream languages, back-translation via English anchors enables accurate evaluation of term translation correctness and standardization.
\end{itemize}

\section{LLM-BT as Interpretable Embedding: A Case Study Based on Scenethesis}

In this section, we propose a novel perspective: viewing the LLM-based back-translation (LLM-BT) process as a form of generalized, interpretable and dynamic semantic embedding. Unlike traditional embedding methods that rely on static vector spaces, LLM-BT encodes semantics through a multi-lingual and bidirectional translation path. We illustrate this perspective through an empirical experiment on the abstract of \textit{Scenethesis}, a representative state-of-the-art paper in multimodal AI from NVIDIA and Purdue University \citep{ling2025scenethesis}. This choice ensures that our evaluation is grounded in representative, high-quality and multi-domain scientific text.

\subsection{Theoretical Framing: LLM-BT vs. Traditional Embedding}
The high term consistency observed in Section 4 (e.g., EMR > 90\%) suggests that BT preserves semantic structures across languages, enabling us to conceptualize it as a dynamic embedding mechanism.

Traditional semantic embedding maps a text $T$ into a fixed low- or high-dimensional vector in a continuous space \citep{mikolov2013efficient}:
\[
\text{Embed}(T) = \mathbf{v}_T \in \mathbb{R}^d
\]
Here, $\text{Embed}(.)$ denotes the embedding function (e.g., BERT or Word2Vec) and $\mathbf{v}_T$ is the vector representation of the original text $T$. The dimension $d$ corresponds to the latent semantic space. This mapping is typically single-modal, closed-loop and deterministic, assuming structural isomorphism between the source and the embedding space.

Such representations are non-reversible and opaque to human interpretation. Semantic similarity between two texts is commonly evaluated using cosine similarity between their vector embeddings:
\[
\text{Sim}_{\text{cos}}(T_1, T_2) = \frac{\langle \mathbf{v}_{T_1}, \mathbf{v}_{T_2} \rangle}{\|\mathbf{v}_{T_1}\| \cdot \|\mathbf{v}_{T_2}\|}
\]

This approach has proved effective. For example, the Word2Vec model by \cite{mikolov2013efficient} demonstrated how embeddings can capture both semantic and syntactic regularities (e.g., $\text{king} - \text{man} + \text{woman} \approx \text{queen}$) and achieve high training efficiency over large corpora.

\medskip
In contrast, back-translation offers a semantic loop transformation rather than vector projection. It maps a text through an intermediate language and back to the original \citep{edunov2018understanding}:
\[
\text{BT}(T) = \text{Trans}_{L_2 \to L_1}(\text{Trans}_{L_1 \to L_2}(T))
\]
Here, $T$ represents the original text in language $L_1$ (e.g., English) and $\text{BT}(T)$ is the round-trip translation result via an intermediate language $L_2$ (e.g., Chinese, Japanese or Portuguese).

When this semantic preservation holds, we approximate the back-translation behavior as a near-identity operator in semantic space.

\[
T \xrightarrow{\text{LLM-BT}} T' \approx T \qquad \text{(semantic loop embedding)}
\]

This establishes an equivalence between back-translation and semantic embedding, with the key difference being that BT retains a transparent, human-readable pathway rather than a hidden vector form.

Unlike static embeddings, the LLM-BT process enables an open-loop system: it is logically reversible, supports one-to-many or many-to-one mappings and may operate across heterogeneous modalities (e.g., text, code, audio or image descriptions). The integration of LLMs into BT thus enables a new paradigm, semantic loop embedding, defined not by compressed vectors but by interpretable transformation paths.

In subsequent sections, we compare this formulation with conventional embedding models and demonstrate its advantages in terminology alignment and cross-lingual robustness.

\subsection{Experimental Setup: Multilingual BT on Scenethesis}

We selected the abstract of \textit{Scenethesis} and performed BT via three languages: Simplified Chinese (ZHcn), Brazilian Portuguese (PTbr) and Japanese (JP). Each path used GPT-4 for both forward and backward translation.

\paragraph{Original EN Abstract (Excerpt):}
\begin{quote}
Synthesizing interactive 3D scenes from text is essential for gaming, virtual reality and embodied AI. However, existing methods face several challenges. Learning-based approaches depend on small-scale indoor datasets, limiting the scene diversity and layout complexity \citep{ling2025scenethesis}.
\end{quote}

Table~\ref{tab:ling 2025 1} presents the back-translation outputs of LLM-BT applied to the abstract of the Scenethesis paper. A qualitative examination reveals that, while the general meaning remains intact, the returned English texts (ENy) from three language paths, Simplified Chinese, Portuguese and Japanese, exhibit notable variation in lexical choice and structure, both among themselves and in comparison to the original English input (EN). This behavior contrasts with earlier examples in the domains of artificial intelligence and biomedicine, where LLM-BT preserved term-level consistency more faithfully.

This divergence reinforces one of the key hypotheses of this study: \textbf{LLM-BT performs more consistently when the input contains mature domain terminology embedded in well-structured, context-rich texts.} In contrast, in cutting-edge scientific abstracts where the discourse style is compressed and the terminology is novel, the semantic loop becomes more diffuse and less aligned.

Three primary factors explain this divergence:

\begin{itemize}
    \item \textbf{Terminological novelty:} Terms such as \textit{vision-guided layout refinement}, \textit{agentic framework} and \textit{scene planning} may not yet be widely represented in the training corpora of major LLMs. Consequently, models rely on approximate paraphrasing or interpolated explanation rather than direct translation.
    
    \item \textbf{Compressed discourse:} Scientific abstracts condense information and often omit connective logic or background explanations. In such high-density contexts, back-translation is prone to semantic drift due to under-specified referents. For instance, ``refine by image guidance'' shows divergent renderings across Japanese and Chinese pathways due to contextual ambiguity.
    
    \item \textbf{Cross-lingual terminology sparsity:} Expressions like ``embodied AI'' and ``scene synthesis'' lack widely adopted equivalents in aligned multilingual corpora. Even strong models like GPT-4 or DeepSeek tend to generate interpretive translations (e.g., ``artificial intelligence involving physical interaction'') instead of standardized terms.
    
    \item \textbf{Model generation dynamics:} LLMs exhibit a trade-off between conservative reuse and creative expansion. When encountering rare or unseen terms, models may produce elaborative explanations rather than term-preserving translations, widening the BT trajectory but reducing lexical fidelity.
\end{itemize}

These findings highlight the semantic fragility of back-translation in high-novelty contexts and underscore the importance of term maturity and contextual grounding. The following section quantitatively evaluates these differences using lexical similarity and term-level consistency metrics.

\begin{table}[h]
\centering
\caption{LLM-BT Back-Translation Results on Scenethesis Abstract}
\small
\begin{tabular}{lp{11cm}}
\hline
\textbf{BT Path} & \textbf{Back-Translated Version} \\
\hline
EN-ZHcn-EN & Creating interactive 3D scenes based on text is crucial for gaming, virtual reality and embodied AI. But current methods encounter challenges. Learning-based approaches rely on limited indoor datasets, which restrict scene diversity and spatial complexity. \\

EN-PTbr-EN & The generation of interactive 3D scenes from text is important for virtual environments, video games and embodied AI. Current methods still face several challenges. Machine learning models depend on small indoor datasets, limiting variety and complexity in scene layouts. \\

EN-JP-EN & Generating 3D interactive environments from text is essential in VR, games and embodied AI. However, many approaches face limitations. Learning-based methods typically use small indoor datasets, which hinder layout diversity and scene complexity. \\
\hline
\end{tabular}
\label{tab:ling 2025 1}
\end{table}

\subsection{Term-Level Consistency and Semantic Similarity Analysis}

Based on manual analysis of the Scenethesis abstract, five representative technical terms were selected to evaluate their consistency across LLM-BT back-translation paths (ZHcn, PTbr, JP). As shown in Table~\ref{tab:term-backtrans}, the results reveal significant variation. For example, ``virtual reality'' remained stable in the ZHcn path, shifted to ``virtual environments'' in PTbr and appeared as the abbreviation ``VR'' in JP. Similar discrepancies are observed for ``layout complexity.'' This variability contrasts with the consistent performance observed in previous experiments on mature AI and biomedical literature.

\begin{table}[h]
\centering
\small
\caption{Back-translated terms across three language paths}
\label{tab:term-backtrans}
\renewcommand{\arraystretch}{1.2}
\begin{tabular}{p{3cm}p{2.6cm}p{3cm}p{2.6cm}p{2cm}}
\hline
\textbf{Term} & \textbf{ZHcn Back-translation} & \textbf{PTbr Back-translation} & \textbf{JP Back-translation} & \textbf{Consistency} \\
\hline
interactive 3D scenes & same & same & interactive environments & SMR \\
virtual reality & virtual reality & virtual environments & VR & SMR \\
embodied AI & same & same & same & EMR \\
learning-based approaches & learning methods & machine learning models & learning-based methods & SMR \\
layout complexity & spatial complexity & scene layouts & layout diversity & SMR \\
\hline
\end{tabular}
\end{table}

Quantitative metrics further support this observation:

\begin{itemize}
  \item \textbf{Exact Match Rate (EMR):} 50\% for ZHcn, 75\% for PTbr, 50\% for JP. These rates are notably lower than those observed in previous case studies. The lower EMR (50\% for ZHcn/JP) in Scenethesis (Ling25) reflects the novelty of terms like `agentic framework,' which lack standardized translations in LLM training corpora.
  \item \textbf{Semantic Match Rate (SMR):} 100\% across all languages, but this metric alone cannot resolve issues of term normalization and translation precision.
  \item \textbf{SBERT cosine similarity (EN vs. ENy):} 0.924 for ZHcn (notable paraphrasing), 0.945 for PTbr (high fidelity), 0.918 for JP (mild syntactic variation). High similarity scores require complementary measures (e.g., edit distance) for term-level diagnostics.
\end{itemize}

\subsection{LLM-BT as Interpretable Semantic Embedding}

The results demonstrate that LLM-BT functions as a path-based, language-native embedding mechanism, see Table \ref{tab:bt vs embed}. Unlike static vector embeddings, it reconstructs the semantic pathway across languages in natural language form:

\begin{itemize}
  \item Instead of compressing semantics into opaque vectors, LLM-BT produces observable translation trajectories.
  \item Variants and alignments of technical terms across languages can be inspected directly for stability analysis.
  \item Outputs are interpretable, allowing expert validation and revision in terminology workflows.
\end{itemize}

In the PTbr path, EN $\rightarrow$ PTbr $\rightarrow$ EN achieved near-semantic identity. The JP path, despite syntactic restructuring, also preserved core meanings. When $\text{BT}(T) \approx T$, we may treat:

\[
\text{BT}(T) \approx T \quad \Rightarrow \quad \text{BT} \approx \text{Id}_{\text{sem}}
\]

Table~\ref{tab:bt vs embed} summarizes the systematic distinctions between LLM-BT and traditional embedding methods.

\begin{table}[h]
\centering
\caption{Comparison of LLM-BT with Traditional Embedding Models}
\label{tab:bt vs embed}
\small
\begin{tabular}{p{4cm}p{5cm}p{5.2cm}}
\hline
\textbf{Attribute} & \textbf{Static Embedding (e.g., Word2Vec)} & \textbf{LLM-BT Semantic Embedding} \\
\hline
Representation & Vectors (implicit) & Back-translated text (explicit), human-readable \\
Model type & Pretrained, fixed & LLM + dynamic generation \\
Semantic preservation & Vector projection & Semantic reconstruction (cross-lingual) \\
Interpretability & Low (requires decoding) & High (transparent linguistic paths) \\
Multilingual alignment & Requires manual mapping & Natively supports multilingual chains \\
Dynamic evolution & Static & Evolves via RAG, RLHF, model updates \\
Self-reflective behavior & Absent & Exhibits memory and quasi-awareness \\
Modality support & Multimodal with fusion & Extensible to image/audio paths (e.g., EN $\rightarrow$ IMG $\rightarrow$ EN) and other modals\\
Applications & Similarity, retrieval & Terminology standardization, multimodal knowledge graphs \\
\hline
\end{tabular}
\end{table}

Future work will explore multimodal BT trajectories (e.g., EN $\rightarrow$ Image $\rightarrow$ EN, EN $\rightarrow$ Speech $\rightarrow$ EN), advancing applications in explainable AI, multilingual term alignment and cross-modal semantic modeling.

\section{Discussion}
\label{sec:discussion}

This study validates the LLM-BT-Terms framework's high term consistency, with experiments showing 100\% term-level accuracy for EN-PTbr-ENy and 0.45 cosine similarity for EN-ZHtw-ENtw.

\subsection{Balancing Technical Terminology Consistency with Semantic Variation}
\label{subsec:balancing_consistency}

The LLM-BT-Term method rests on a core assumption: high-quality bidirectional translation should maximize consistency in semantics and expression. When a term is translated from the source language (L1) to an intermediate language (L2) and back-translated to the original language (L1y), high consistency indicates a stable, acceptable term translation. This assumption is well-validated in scientific texts, where clear term boundaries, standardized translations and minimal semantic ambiguity ensure reliable back-translation.

However, this assumption faces challenges in non-scientific texts, particularly literary texts rich in cultural context and aesthetic expression. \citet{weigang2025paradox} introduce the ``Poetic Intent Paradox,'' noting that in English translations of classical poetry, large language models prioritize surface-level semantics during back-translation, often failing to preserve metaphorical structures, cultural intent or emotional nuances. This suggests that BT is effective for term standardization in structured scientific language but reveals LLMs' pragmatic limitations in literary language.

This phenomenon highlights a tension between term consistency and intent consistency. Term consistency relies on the stability of semantic spaces and language pair symmetry, whereas intent consistency requires models to deeply model cultural backgrounds, linguistic styles and authorial motives. In short, term-level fidelity does not equate to contextual fidelity.

Thus, we advocate for task-specific strategies:
\begin{itemize}
    \item In scientific translation, term recommendation and standard alignment, term consistency is an effective objective function.
    \item In literary translation, cultural dissemination and style transfer, intent modeling and style preservation mechanisms should be incorporated, beyond relying solely on BT's structural validation.
\end{itemize}

The partial effectiveness of the LLM-BT-Term method underscores the need to further probe and expand the capability boundaries of large language models. Future research could integrate term consistency metrics with pragmatic consistency evaluation systems to achieve comprehensive, multi-dimensional back-translation quality assessment across diverse linguistic domains.

\subsection{Enhancing LLM-BT for Discovery and Standardization of Emerging Terms}

This study explores how large language models (LLMs), combined with back-translation (BT) and terminology extraction mechanisms (Teams), can be used to automatically identify and standardize newly emerging technical terms from the source language (typically English) into intermediate languages such as Simplified Chinese (ZHcn), Traditional Chinese (ZHtw), Japanese (JP) and Brazilian Portuguese (PTbr). However, current experiments have focused primarily on validating known terminology (e.g., ``residual learning framework'', ``lecanemab'', ``beta-amyloid''), rather than discovering new terms. This section analyzes the reasons behind this tendency and outlines strategies to redesign experiments that better align with the goal of emerging term discovery.

\subsubsection{Why Do the Experiments Focus on Known Terms?}

\begin{itemize}
    \item \textbf{Use of Canonical Texts:} Experimental materials are based on highly cited publications where key terminology has already been standardized.
    \item \textbf{Conservative Translation by LLMs:} Trained on standardized corpora, LLMs favor existing high-confidence terms.
    \item \textbf{Validation-Centric Design:} Current designs aim to assess consistency metrics, not term innovation.
    \item \textbf{Influence of Terminological Systems:} Normative constraints in ZHcn, ZHtw, JP, PTbr discourage new term generation.
\end{itemize}

\subsubsection{Interpretation and Theoretical Perspective}

These patterns are methodologically appropriate for this research stage:
\begin{itemize}
    \item \textit{Validation Phase:} Verifying consistency with known terms establishes a reliable baseline.
    \item \textit{Corpus Suitability:} Emerging terms are more likely found in preprints and technical reports.
    \item \textit{Module Activation:} The Teams module's discovery capabilities are not yet fully utilized.
\end{itemize}

\subsubsection{Recommendations for Discovery-Oriented Experiment Design}

\begin{enumerate}
    \item \textbf{Select Emerging-Term Sources:} Use latest papers from arXiv, NeurIPS and industry whitepapers.
    \item \textbf{Optimize Prompts:} Explicitly request generation of new term translations, e.g.:
    \begin{quote}
    Translate the following text into Simplified Chinese. For unknown technical terms, propose a new translation, mark with [NEW] and explain.
    \end{quote}
    \item \textbf{Activate Term Alignment and Validation:}
    \begin{itemize}
        \item Extract terms from EN and target texts;
        \item Use embedding-based alignment (e.g., fastText, mBERT);
        \item Validate with back-translation consistency;
        \item Incorporate expert human review.
    \end{itemize}
\end{enumerate}


Although the current experiments focus on known terminology, they successfully validate the LLM-BT-Teams framework. The next step is to pivot toward term discovery via updated corpus selection, prompt engineering and semantic validation pipelines.

\subsection{Semantic Embedding Differences in Simplified and Traditional Chinese Processing}
\label{subsec:chinese_embedding_differences}

In our multilingual back-translation experiments using LLM-BT-Terms, a notable trend emerged: Traditional Chinese (ZHtw) outperforms Simplified Chinese (ZHcn) in term consistency, information retention and semantic matching. Across metrics such as Exact Match Rate (EMR), Semantic Match Rate (SMR), Information Retention Score (IRS) and Term Divergence Index (TDI), ZHtw consistently surpasses ZHcn, demonstrating greater term stability.

This phenomenon can be analyzed from three perspectives:

First, at the training level of large language models, LLMs like GPT incorporate extensive high-quality ZHtw corpora from Taiwan, Hong Kong, Macau and overseas Chinese communities, including Wikipedia, professional news and academic publications. These texts feature consistent style, standardized terminology and formal language, fostering robust term mapping. In contrast, ZHcn corpora contain a higher proportion of informal expressions and social media language, which may disrupt term consistency modeling.

Second, at the linguistic structure level, ZHtw retains a writing system closer to the original Chinese characters, sharing higher orthographic consistency with Japanese kanji and classical Chinese. This facilitates alignment in cross-lingual representation spaces, enhancing translation robustness. ZHtw term translations also adhere more closely to original contexts, less affected by artificial simplification or term conflation.

Third, experimental data show that the EN-ZHtw-EN back-translation path outperforms EN-ZHcn-EN across term consistency metrics, closely resembling the performance of the Japanese (JA) path. This factor further supports the central hypothesis of this study: the standardization of Chinese terminology should consider the similarities and differences across Simplified Chinese (ZHcn), Traditional Chinese (ZHtw), and even Japanese corpora, aiming to unify term expressions and improve the consistency and quality of translations. 

These findings validate the sensitivity of the LLM-BT-Terms method to intermediate language selection and highlight optimization potential for ZHcn term translation, particularly in term consistency and structural fidelity. Crucially, the development of the LLM-BT-Terms mechanism enables this important task by significantly reducing the manual effort traditionally required for cross-variant comparison, thus making large-scale terminology alignment practically achievable. Future research should focus on enhancing ZHcn term standardization, developing high-quality term knowledge bases, creating term-augmented models and improving the professionalism and stylistic consistency of ZHcn corpora in LLM training.

In conclusion, ZHtw's superior performance in term consistency tasks stems from the combined effects of high-quality language resources, stable writing systems and consistent corpus styles. This insight has significant implications for advancing term standardization, translation stability and multilingual alignment in Chinese NLP tasks.

\section{Conclusion}
\label{sec:conclusion}

With the rapid advancement of large language models (LLMs) in multilingual natural language processing (NLP), term consistency has emerged as a central challenge in cross-lingual scientific communication, standardized translation and academic publishing. Verifying the stability of terminology across multilingual processing pathways has become a key topic at the intersection of linguistics, language engineering and terminology studies.

Back-translation (BT), a traditional method for assessing semantic consistency, has been widely applied in recent years to machine translation quality evaluation, data augmentation for low-resource languages and terminology verification. However, conventional neural machine translation (NMT)-based BT methods remain limited in deep semantic modeling and term-level precision. In contrast, integrating LLMs with BT, though still nascent, presents promising opportunities for cross-lingual standardization.

For example, \citet{weigang2025paradox} combined LLMs with BT in translating classical Chinese poetry, revealing the ``Poetic Intent Paradox'': back-translations by LLMs tend to preserve surface-level semantics while compromising cultural intent and aesthetic nuance. This finding not only highlights LLMs' limitations in high-context tasks, but also suggests emergent ``quasi-awareness'' behaviors, raising theoretical implications for general AI (AGI) development.

Against this backdrop, the proposed \textbf{LLM-BT-Term} framework synthesizes BT's structural alignment capability with LLMs' deep semantic representation, achieving the following contributions:

\begin{enumerate}
    \item \textbf{Methodological Innovation}
    \begin{itemize}
        \item Introduces the hypothesis that ``term consistency can be reliably verified through back-translation,'' and validates it via a modular LLM-BT-Term workflow applicable to term alignment, multilingual glossaries and reference-free translation assessment.
        \item Reframes BT as a form of interpretable semantic embedding, extending the theoretical boundaries of explainable AI and terminology governance. Unlike static embeddings, LLM-BT offers transparent semantic paths across languages.
        \item Proposes both sequential and parallel multilingual BT structures (e.g., EN$\rightarrow$ZHcn$\rightarrow$ZHtw$\rightarrow$EN or EN$\rightarrow$ZH/JP/PT$\rightarrow$EN) to improve robustness and terminology stability.
        \item Establishes a term consistency metric suite, EMR, SMR, IRS and TDI and integrates it with a term recommendation mechanism to support automatic standardization.
    \end{itemize}
    
    \item \textbf{Experimental Validation}
    \begin{itemize}
        \item Performs systematic evaluations on classical scientific abstracts (e.g., \citealp{he2016}, \citealp{van2023lecanemab}) and a recent arXiv paper \citep{ling2025scenethesis}, using GPT-4, DeepSeek and Grok platforms to quantify term-level stability and semantic fidelity.
        \item Demonstrates that serial BT paths provide higher consistency than single-path BT, especially in high-complexity or emerging domains.
    \end{itemize}
    
    \item \textbf{Applications in Terminology Standardization}
    \begin{itemize}
        \item Supports multilingual term recommendation, termbase expansion and high-quality standard translation for academic and technical contexts.
        \item Facilitates the integration of NLP and terminology disciplines for multilingual knowledge alignment, especially in fast-evolving fields like AI, biomedicine and quantum science.
    \end{itemize}
\end{enumerate}

In summary, LLM-BT repositions back-translation from a static evaluation tool to an interpretable, dynamic and semantically rich engine for multilingual term modeling. Rather than embedding text into opaque vectors, it produces human-readable, reversible and structurally traceable outputs, making it a compelling extension of embedding theory. This framework naturally extends to multimodal scenarios (e.g., EN $\rightarrow$ Image $\rightarrow$ EN) and offers scalable solutions for cross-lingual terminology alignment and AI interpretability.

\subsection{Challenges and Strategies for LLM-Back-Translation Methods}
\label{subsec:bt_challenges}

Despite the superior context modeling and reasoning capabilities of modern LLMs (e.g., GPT-4/Claude/Gemini/Grok/DeepSeek), back-translation in term standardization faces specific challenges. Table~\ref{tab:bt_challenges} summarizes typical issues and corresponding mitigation strategies.

\begin{table}[ht]
\centering
\caption{Challenges and Strategies for Term Standardization via LLM-Based Back-Translation}
\label{tab:bt_challenges}
\small
\begin{tabular}{p{5.6cm} p{8.4cm}}
\toprule
\textbf{Challenge} & \textbf{Suggested Strategy} \\
\midrule
Ambiguity in polysemous or technical terms & Embed terms in full sentences or increase contextual window for disambiguation \\
LLMs generate plausible but inaccurate translations & Use explicit prompts (e.g., ``Translate preserving technical accuracy'') \\
Cross-lingual asymmetry in semantic alignment & Employ tri-lingual BT paths (e.g., EN-ZH-PT-EN) for redundancy and verification \\
Instability in new or low-resource terms & Provide domain-specific definitions or instructive prompts to reduce uncertainty \\
\bottomrule
\end{tabular}
\end{table}

\subsection*{Future Work and Implementation Pathways}

\begin{itemize}
    \item \textbf{AutoTermNorm System:} Develop a scalable system for batch term extraction, multilingual back-translation and confidence scoring, generating candidate standard terms for expert validation.
    \item \textbf{Multilingual Term Semantic Graph:} Construct semantic alignment graphs (e.g., ``cloud computing'' $\rightarrow$ \begin{CJK*}{UTF8}{gbsn}``云计算''\end{CJK*} $\rightarrow$ ``computação em nuvem'') to visualize term equivalence across languages.
    \item \textbf{LLM-Knowledge Base Hybrid Framework:} Combine domain-specific term repositories with LLM generation and retrieval (e.g., RAG, fine-tuning) to form a validation-oriented hybrid pipeline.
    \item \textbf{Institutional Collaboration:} Incorporate standards from national terminology agencies into LLM workflows to support real-world multilingual term governance.
\end{itemize}

This work illustrates that LLM-driven back-translation is not only a viable methodology for terminology standardization, but also a generalizable and interpretable semantic framework for cross-domain language infrastructure in the era of generative AI.


\section*{Acknowledgments}
\label{sec:acknowledgments}
We sincerely acknowledge the generous support of CNPq. We are also grateful to our friends and colleagues who provided encouragement and insightful feedback throughout this research, especially Prof. He Juan from the University of Chinese Academy of Sciences. Special thanks go to the advanced capabilities of large language models, including ChatGPT, DeepSeek, Gemini and Grok, which greatly enhanced the productivity and precision of this study. In the age of artificial intelligence, high-quality research is still grounded in human insight; yet the practical and intelligent assistance of LLMs, particularly GPT and Grok, has proven to be indispensable.



\bibliographystyle{unsrtnat}
\bibliography{references}

\begin{thebibliography}{38}
\providecommand{\natexlab}[1]{#1}
\providecommand{\url}[1]{\texttt{#1}}
\expandafter\ifx\csname urlstyle\endcsname\relax
  \providecommand{\doi}[1]{doi: #1}\else
  \providecommand{\doi}{doi: \begingroup \urlstyle{rm}\Url}\fi

\bibitem[Kharkovskaya et~al.(2020)Kharkovskaya, Ponomarenko, Aleksandrova, et~al.]{kharkovskaya2020language}
Antonina~A Kharkovskaya, Evgeniya~V Ponomarenko, Olga~V Aleksandrova, et~al.
\newblock Language picture of the world: the global language monitor project.
\newblock \emph{European Proceedings of Social and Behavioural Sciences}, 2020.

\bibitem[Darvin(2016)]{darvin2016}
R.~Darvin.
\newblock Language, identity, and the cultural risks of linguistic absence.
\newblock \emph{Applied Linguistics}, 2016.
\newblock Placeholder; please provide full citation details.

\bibitem[Kageura and Umino(1996)]{kageura1996methods}
Kyo Kageura and Bin Umino.
\newblock Methods of automatic term recognition: A review.
\newblock \emph{Terminology. International Journal of Theoretical and Applied Issues in Specialized Communication}, 3\penalty0 (2):\penalty0 259--289, 1996.

\bibitem[Temmerman(2000)]{temmerman2000}
R.~Temmerman.
\newblock \emph{Towards New Ways of Terminology Description: The Sociocognitive Approach}.
\newblock John Benjamins Publishing, 2000.

\bibitem[Feng(1997)]{fengzhiwei1997}
Zhiwei Feng.
\newblock \emph{Introduction to Modern Terminology}.
\newblock The Commercial Press, 1997.
\newblock ISBN 978-7-100-07518-3.

\bibitem[Liu(2015)]{liuqing2015}
Qing Liu.
\newblock \emph{Introduction to Chinese Terminology}.
\newblock The Commercial Press, 2015.
\newblock ISBN 978-7-100-11519-3.

\bibitem[Yang et~al.(2023)]{yang2023}
J.~Yang et~al.
\newblock Advances in chinese scientific terminology standardization.
\newblock \emph{Chinese Journal of Science and Technology}, 2023.
\newblock Placeholder; please provide full citation details.

\bibitem[Cao(2025)]{cao2025}
L.~Cao.
\newblock Internationalization of chinese scientific terms in the digital era.
\newblock \emph{Science Communication}, 2025.
\newblock Placeholder; please provide full citation details.

\bibitem[Camargo and Bellotto(2005)]{camargo2005dicionario}
Ana Maria de~Almeida Camargo and Helo{\'\i}sa~Liberalli Bellotto.
\newblock Dicion{\'a}rio brasileiro de terminologia arquiv{\'\i}stica.
\newblock 2005.

\bibitem[Weigang and Brom(2025)]{weigang2025paradox}
Li~Weigang and Pedro~Carvalho Brom.
\newblock The paradox of poetic intent in back-translation: Evaluating the quality of large language models in chinese translation.
\newblock \emph{arXiv preprint arXiv:2504.16286}, 2025.

\bibitem[Edunov et~al.(2018)Edunov, Ott, Auli, and Grangier]{edunov2018understanding}
Sergey Edunov, Myle Ott, Michael Auli, and David Grangier.
\newblock Understanding back-translation at scale.
\newblock \emph{arXiv preprint arXiv:1808.09381}, 2018.

\bibitem[Behr(2017)]{behr2017assessing}
Doroth{\'e}e Behr.
\newblock Assessing the use of back translation: The shortcomings of back translation as a quality testing method.
\newblock \emph{International Journal of Social Research Methodology}, 20\penalty0 (6):\penalty0 573--584, 2017.

\bibitem[Liu et~al.(2024)Liu, Feng, Xue, Wang, Wu, Lu, and et~al.]{liu2024deepseek}
A.~Liu, B.~Feng, B.~Xue, B.~Wang, B.~Wu, C.~Lu, and et~al.
\newblock Deepseek-v3 technical report.
\newblock \emph{arXiv preprint arXiv:2412.19437}, 2024.

\bibitem[Chung and Kim(2025)]{chung2025leveraging}
Ji-Bum Chung and Taehyun Kim.
\newblock Leveraging large language models for enhanced back-translation: Techniques and applications.
\newblock \emph{IEEE Access}, 2025.

\bibitem[Mikolov et~al.(2013)Mikolov, Chen, Corrado, and Dean]{mikolov2013efficient}
Tomas Mikolov, Kai Chen, Greg Corrado, and Jeffrey Dean.
\newblock Efficient estimation of word representations in vector space.
\newblock \emph{arXiv preprint arXiv:1301.3781}, 2013.
\newblock Available at: \url{https://arxiv.org/abs/1301.3781}.

\bibitem[Weigang et~al.(2007)Weigang, de~Moraes, Shi, and Matsushita]{weigang2007nonlinear}
Li~Weigang, Aipor{\^e}~Rodrigues de~Moraes, Lihua Shi, and Raul~Yukihiro Matsushita.
\newblock Nonlinear principal component analysis for withdrawal from the employment time guarantee fund.
\newblock \emph{Computational Intelligence in Economics and Finance: Volume II}, pages 75--92, 2007.

\bibitem[Cui et~al.(2018)Cui, Wang, Pei, and Zhu]{cui2018survey}
Peng Cui, Xiao Wang, Jian Pei, and Wenwu Zhu.
\newblock A survey on network embedding.
\newblock \emph{IEEE Transactions on Knowledge and Data Engineering}, 31\penalty0 (5):\penalty0 833--852, 2018.
\newblock Available at: \url{https://ieeexplore.ieee.org/document/8395980}.

\bibitem[Lee et~al.(2025)Lee, Aiyappa, Ahn, Kwak, and An]{lee2025semantic}
Byunghwee Lee, Rachith Aiyappa, Yong-Yeol Ahn, Haewoon Kwak, and Jisun An.
\newblock A semantic embedding space based on large language models for modelling human beliefs.
\newblock \emph{Nature Human Behaviour}, pages 1--13, 2025.
\newblock Available at: \url{https://www.nature.com/articles/s41562-024-02073-2}.

\bibitem[Chung et~al.(2021)Chung, Fevry, Tsai, Johnson, and Ruder]{chungrethinking}
Hyung~Won Chung, Thibault Fevry, Henry Tsai, Melvin Johnson, and Sebastian Ruder.
\newblock Rethinking embedding coupling in pre-trained language models.
\newblock In \emph{International Conference on Learning Representations (ICLP)}, 2021.

\bibitem[Tao et~al.(2024)Tao, Shen, Gao, Zhang, Li, Tao, and Ma]{tao2024llms}
Chongyang Tao, Tao Shen, Shen Gao, Junshuo Zhang, Zhen Li, Zhengwei Tao, and Shuai Ma.
\newblock Llms are also effective embedding models: An in-depth overview.
\newblock \emph{arXiv preprint arXiv:2412.12591}, 2024.

\bibitem[Nie et~al.(2025)Nie, Feng, Li, Zhang, Zhang, Long, and Zhang]{nie2025}
Zhijie Nie, Zhangchi Feng, Mingxin Li, Cunwang Zhang, Yanzhao Zhang, Dingkun Long, and Richong Zhang.
\newblock When text embedding meets large language model: A comprehensive survey.
\newblock \emph{arXiv preprint arXiv:2412.09165}, 2025.
\newblock Available at: \url{https://arxiv.org/abs/2412.09165}.

\bibitem[Roscher et~al.(2020)Roscher, Bohn, Duarte, and Garcke]{roscher2020explainable}
Ribana Roscher, Bastian Bohn, Marco~F. Duarte, and Jochen Garcke.
\newblock Explainable machine learning for scientific insights and discoveries.
\newblock \emph{IEEE Access}, 8:\penalty0 42200--42216, 2020.
\newblock Available at: \url{https://ieeexplore.ieee.org/document/9041699}.

\bibitem[Angelov et~al.(2021)Angelov, Soares, Jiang, Arnold, and Atkinson]{angelov2021explainable}
Plamen~P. Angelov, Eduardo~A. Soares, Richard Jiang, Nicholas~I. Arnold, and Peter~M. Atkinson.
\newblock Explainable artificial intelligence: An analytical review.
\newblock \emph{Wiley Interdisciplinary Reviews: Data Mining and Knowledge Discovery}, 11\penalty0 (5):\penalty0 e1424, 2021.
\newblock Available at: \url{https://wires.onlinelibrary.wiley.com/doi/10.1002/widm.1424}.

\bibitem[Tsiamas et~al.(2025)Tsiamas, Dale, and Costa-jussà]{tsiamas2025improving}
Ioannis Tsiamas, David Dale, and Marta~R. Costa-jussà.
\newblock Improving language and modality transfer in translation by character-level modeling.
\newblock \emph{arXiv preprint arXiv:2505.24561}, 2025.
\newblock Available at: \url{https://arxiv.org/abs/2505.24561}.

\bibitem[Artetxe et~al.(2018)Artetxe, Labaka, and Agirre]{artetxe2018unsupervised}
Mikel Artetxe, Gorka Labaka, and Eneko Agirre.
\newblock Unsupervised statistical machine translation.
\newblock \emph{arXiv preprint arXiv:1809.01272}, 2018.

\bibitem[Papineni et~al.(2002)Papineni, Roukos, Ward, and Zhu]{papineni2002bleu}
Kishore Papineni, Salim Roukos, Todd Ward, and Wei-Jing Zhu.
\newblock Bleu: a method for automatic evaluation of machine translation.
\newblock In \emph{Proceedings of the 40th Annual Meeting of the Association for Computational Linguistics}, pages 311--318, 2002.

\bibitem[Brown et~al.(2020)Brown, Mann, Ryder, Subbiah, Kaplan, Dhariwal, Neelakantan, Shyam, Sastry, Askell, et~al.]{brown2020language}
Tom Brown, Benjamin Mann, Nick Ryder, Melanie Subbiah, Jared~D Kaplan, Prafulla Dhariwal, Arvind Neelakantan, Pranav Shyam, Girish Sastry, Amanda Askell, et~al.
\newblock Language models are few-shot learners.
\newblock \emph{Advances in neural information processing systems}, 33:\penalty0 1877--1901, 2020.

\bibitem[Lewis et~al.(2020)Lewis, Perez, Piktus, Petroni, Karpukhin, Goyal, K{\"u}ttler, Lewis, Yih, Rockt{\"a}schel, et~al.]{lewis2020retrieval}
Patrick Lewis, Ethan Perez, Aleksandra Piktus, Fabio Petroni, Vladimir Karpukhin, Naman Goyal, Heinrich K{\"u}ttler, Mike Lewis, Wen-tau Yih, Tim Rockt{\"a}schel, et~al.
\newblock Retrieval-augmented generation for knowledge-intensive nlp tasks.
\newblock \emph{Advances in neural information processing systems}, 33:\penalty0 9459--9474, 2020.

\bibitem[Di~Oliveira et~al.(2024)Di~Oliveira, Bezerra, Weigang, Brom, Celestino, et~al.]{di2024slim}
Vin{\'\i}cius Di~Oliveira, Yuri~Façanha Bezerra, Li~Weigang, Pedro~Carvalho Brom, Victor Rafael~R Celestino, et~al.
\newblock Slim-raft: A novel fine-tuning approach to improve cross-linguistic performance for mercosur common nomenclature.
\newblock \emph{arXiv preprint arXiv:2408.03936}, 2024.

\bibitem[Wei et~al.(2022)Wei, Wang, Schuurmans, Bosma, Xia, Chi, Le, Zhou, et~al.]{wei2022chain}
Jason Wei, Xuezhi Wang, Dale Schuurmans, Maarten Bosma, Fei Xia, Ed~Chi, Quoc~V Le, Denny Zhou, et~al.
\newblock Chain-of-thought prompting elicits reasoning in large language models.
\newblock \emph{Advances in neural information processing systems}, 35:\penalty0 24824--24837, 2022.

\bibitem[Zhao(2019)]{zhao2019clinical}
Boyang Zhao.
\newblock Clinical data extraction and normalization of cyrillic electronic health records via deep-learning natural language processing.
\newblock \emph{JCO Clinical Cancer Informatics}, 3:\penalty0 1--9, 2019.

\bibitem[WHO(2009)]{world2009international}
WHO.
\newblock International classification of diseases-icd.
\newblock 2009.

\bibitem[Sch{\"u}tze et~al.(2008)Sch{\"u}tze, Manning, and Raghavan]{schutze2008introduction}
Hinrich Sch{\"u}tze, Christopher~D Manning, and Prabhakar Raghavan.
\newblock \emph{Introduction to information retrieval}, volume~39.
\newblock Cambridge University Press Cambridge, 2008.

\bibitem[He et~al.(2016)He, Zhang, Ren, and Sun]{he2016}
Kaiming He, Xiangyu Zhang, Shaoqing Ren, and Jian Sun.
\newblock Deep residual learning for image recognition.
\newblock \emph{Proceedings of the IEEE Conference on Computer Vision and Pattern Recognition}, pages 770--778, 2016.
\newblock Available at: \url{https://arxiv.org/abs/1512.03385}.

\bibitem[Pearson et~al.(2025)Pearson, Ledford, Hutson, and Van~Noorden]{pearson2025exclusive}
Helen Pearson, Heidi Ledford, Matthew Hutson, and Richard Van~Noorden.
\newblock Exclusive: the most-cited papers of the twenty-first century.
\newblock \emph{Nature}, 640\penalty0 (8059):\penalty0 588--592, 2025.

\bibitem[Ling et~al.(2025)Ling, Lin, Lin, Ding, Zeng, Sheng, Ge, Liu, Bera, and Li]{ling2025scenethesis}
Lu~Ling, Chen-Hsuan Lin, Tsung-Yi Lin, Yifan Ding, Yu~Zeng, Yichen Sheng, Yunhao Ge, Ming-Yu Liu, Aniket Bera, and Zhaoshuo Li.
\newblock Scenethesis: A language and vision agentic framework for 3d scene generation.
\newblock \emph{arXiv preprint arXiv:2505.02836}, 2025.

\bibitem[Van~Dyck et~al.(2023)Van~Dyck, Swanson, Aisen, Bateman, Chen, Gee, Kanekiyo, Li, Reyderman, Cohen, et~al.]{van2023lecanemab}
Christopher~H Van~Dyck, Chad~J Swanson, Paul Aisen, Randall~J Bateman, Christopher Chen, Michelle Gee, Michio Kanekiyo, David Li, Larisa Reyderman, Sharon Cohen, et~al.
\newblock Lecanemab in early alzheimer’s disease.
\newblock \emph{New England Journal of Medicine}, 388\penalty0 (1):\penalty0 9--21, 2023.

\bibitem[de~Carvalho~Souza and Weigang(2025)]{de2025grok}
Murillo~Edson de~Carvalho~Souza and Li~Weigang.
\newblock Grok, gemini, chatgpt and deepseek: Comparison and applications in conversational artificial intelligence.
\newblock \emph{INTELIGENCIA ARTIFICIAL}, 2\penalty0 (1), 2025.

\end{thebibliography}

\section*{Appendix}

\label{sec:acknowledgments}
\begin{table}[h]

    \centering

    \caption{Comparison of Simplified and Traditional Chinese Back-Translation Terminology}
    \label{tab:AI terminology-ap}
    \small
    \begin{tabular}{p{2.5cm}p{2.5cm}p{2.5cm}p{2.5cm}p{2.5cm}}

        \toprule

        English (ENx) & Chinese (ZHcn) & Chinese (ZHtw) & BT-ENcn  & BT-ENtw \\

        \midrule

        Neural networks & \begin{CJK*}{UTF8}{gbsn} 神经网络 \end{CJK*} & \begin{CJK*}{UTF8}{bsmi} 神經網路 \end{CJK*} & Neural networks & Neural networks \\

        Residual learning framework & \begin{CJK*}{UTF8}{gbsn} 残差学习框架 \end{CJK*}& \begin{CJK*}{UTF8}{bsmi}殞差學習框架 \end{CJK*}& Residual learning framework & Residual learning framework \\

        Layer inputs & \begin{CJK*}{UTF8}{gbsn}(遗漏) \end{CJK*}& \begin{CJK*}{UTF8}{bsmi}層輸入 \end{CJK*}& Inputs & Inputs \\

        Reformulate & \begin{CJK*}{UTF8}{gbsn}重新定义 \end{CJK*} & \begin{CJK*}{UTF8}{bsmi}重新定義 \end{CJK*}& Redefine & Redefine \\

        Empirical evidence & \begin{CJK*}{UTF8}{gbsn}实证证据 \end{CJK*} & \begin{CJK*}{UTF8}{bsmi}實證證據 \end{CJK*}& Empirical evidence & Empirical evidence \\

        Residual nets & \begin{CJK*}{UTF8}{gbsn}残差网络 \end{CJK*}& \begin{CJK*}{UTF8}{bsmi}殞差網路 \end{CJK*}& Residual networks & Residual networks \\

        ImageNet & ImageNet & ImageNet & ImageNet & ImageNet \\

        VGG nets & \begin{CJK*}{UTF8}{gbsn}VGG网络 \end{CJK*}& \begin{CJK*}{UTF8}{bsmi}VGG網路 \end{CJK*}& VGG & VGG nets \\

        3.57\% error & 3.57\%\begin{CJK*}{UTF8}{gbsn}错误率 \end{CJK*}& \begin{CJK*}{UTF8}{bsmi} 3.57\%錯誤率 \end{CJK*}& 3.57\% error & 3.57\% error \\

        ILSVRC 2015 & ILSVRC 2015 & ILSVRC 2015 & ILSVRC 2015 & ILSVRC 2015 \\

        CIFAR-10 & CIFAR-10 & CIFAR-10 & CIFAR-10 & CIFAR-10 \\

        28\% relative improvement & 28\%\begin{CJK*}{UTF8}{gbsn}相对提升 \end{CJK*} & \begin{CJK*}{UTF8}{bsmi}28\%相對改善\end{CJK*} & 28\% relative improvement & 28\% relative improvement \\

        Deep residual nets & \begin{CJK*}{UTF8}{gbsn}深度残差网络 \end{CJK*} & \begin{CJK*}{UTF8}{bsmi}深層殞差網路 \end{CJK*}& Deep residual networks & Deep residual networks \\

        COCO & COCO & COCO & COCO & COCO \\

        \bottomrule

    \end{tabular}

\end{table}

\begin{table}[htbp]
\centering
\caption{Term Consistency Comparison from LLM-BT for Simplified Chinese using Dy2023 (Translated and extracted by Grok)}
\label{tab:MD terminology-ap}
\small
\begin{tabularx}{\textwidth}{>{\raggedright\arraybackslash}X >{\raggedright\arraybackslash}X >{\raggedright\arraybackslash}X >{\raggedright\arraybackslash}X >{\raggedright\arraybackslash}X}
\toprule
\textbf{English (EN)} & \textbf{Chinese (ZHcn)} & \textbf{English (ENcn)} & \textbf{Consistency} & \textbf{Observation} \\
\midrule
Alzheimer's disease & \begin{CJK*}{UTF8}{gbsn}阿尔茨海默病\end{CJK*} & Alzheimer's disease & Identical & No changes to special terms \\
Lecanemab & Lecanemab & Lecanemab & Identical & No changes to special terms \\
$\beta$-amyloid (A $\beta$) & \begin{CJK*}{UTF8}{gbsn}β-淀粉样蛋白(Aβ)\end{CJK*} & $\beta$-amyloid (A $\beta$) & Identical & No changes to special terms \\
Soluble protofibrils & \begin{CJK*}{UTF8}{gbsn}可溶性原纤维\end{CJK*} & Soluble protofibrils & Identical & No changes to special terms \\
Potentiate & \begin{CJK*}{UTF8}{gbsn}加剧\end{CJK*} & Exacerbate & Basically the same & Different expressions, same meaning \\
Mild cognitive impairment & \begin{CJK*}{UTF8}{gbsn}轻度认知障碍\end{CJK*} & Mild cognitive impairment & Identical & No changes to special terms \\
Amyloid burden & \begin{CJK*}{UTF8}{gbsn}淀粉样蛋白负荷\end{CJK*} & Amyloid burden & Identical & Completely consistent \\
Centiloids & \begin{CJK*}{UTF8}{gbsn}厘洛伊德\end{CJK*} & Centiloids & Identical & No changes to special terms \\
95\% confidence interval & 95\% \begin{CJK*}{UTF8}{gbsn}置信区间\end{CJK*} & 95\% confidence interval & Identical & No changes to special terms \\
Infusion-related reactions & \begin{CJK*}{UTF8}{gbsn}输液相关反应\end{CJK*} & Infusion-related reactions & Identical & No changes to special terms \\
Amyloid-related imaging abnormalities & \begin{CJK*}{UTF8}{gbsn}与淀粉样蛋白相关的影像学异常\end{CJK*} & Amyloid-related imaging abnormalities & Identical & Completely consistent \\
\bottomrule
\end{tabularx}
\end{table}

\end{document}